\documentclass[runningheads,a4paper,draft]{svjour3}

\usepackage[utf8]{inputenc}
\usepackage{lmodern}
\usepackage{todonotes}

\usepackage{alltt} 
\usepackage{amsfonts}
\usepackage{booktabs}

\usepackage{tikz}
\usetikzlibrary{shapes,arrows}
\usepackage{pgfplots}
\usepackage{xcolor}
\usepackage{url}
\usepackage{subfigure}

\usepackage{xspace}
\usepackage{amsmath}

\DeclareMathOperator*{\argmax}{\arg\!\max}

\def\holfour{\textsf{HOL4}\xspace}
\def\isabelle{\textsf{Isabelle}\xspace}
\def\hollight{\textsf{HOL Light}\xspace}
\def\mizar{\textsf{Mizar}\xspace}
\def\coq{\textsf{Coq}\xspace}

\def\eprover{\textsf{E~prover}\xspace}

\def\sml{\textsf{SML}\xspace}

\def\holyhammer{\textsf{HOL(y)Hammer}\xspace}
\def\sledgehammer{\textsf{Sledgehammer}\xspace}

\def\ltac{\textsf{Ltac}\xspace}

\def\metis{\textsf{Metis}\xspace}

\def\tactictoe{\textsf{TacticToe}\xspace}

\newcommand{\ra}[1]{\renewcommand{\arraystretch}{#1}}
\usepackage{fancyvrb}
\usepackage{tikz} 
\usetikzlibrary{arrows,shapes}
\usetikzlibrary{trees,positioning,fit}
\tikzstyle{block} =
  [rectangle, draw,
   minimum width=9em, text centered, rounded corners, minimum height=3.5em]
\tikzstyle{line}=[draw]
\tikzstyle{cloud} =
  [draw, text centered, ellipse, minimum height=3.5em, minimum width=9em]

\usetikzlibrary{arrows,shapes,calc}
\usetikzlibrary{trees,positioning,fit}
\tikzstyle{arrow}=[draw,-to,thick]
\tikzstyle{bluearrow}=[draw,-to,thick,blue]
\tikzstyle{tcircle} = [circle, draw, fill=white]
\tikzstyle{tsquare} = [rectangle, draw, fill=white]
\tikzstyle{bcircle} = [circle, draw, blue, fill=blue]
\tikzstyle{gcircle} = [circle, draw, mygreen, fill=mygreen]

\title{TacticToe: Learning to Prove with Tactics}

\author{\mbox{Thibault Gauthier} \and \mbox{Cezary Kaliszyk} \and \mbox{Josef
Urban} \and \mbox{Ramana Kumar} \and \mbox{Michael Norrish}}

\authorrunning{Gauthier et al.}

\institute{Thibault Gauthier and Cezary Kaliszyk \at
Department of Computer Science, University of Innsbruck,
Innsbruck, Austria\\ \url{{thibault.gauthier,cezary.kaliszyk}@uibk.ac.at}
\and
Josef Urban \at Czech Technical University, Prague\\\url{josef.urban@gmail.com}
\and Ramana Kumar and Michael Norrish \at Data61}

\usepackage[final]{listings}
\lstdefinelanguage{SML}%
{columns=fullflexible,
  keywords={THEN,THENL,val,open,store_thm,fun,fn,let,in,end},%
  frame=none,
  sensitive=true,
  keywordstyle=\fontfamily{lmss}\footnotesize\selectfont,%
  basicstyle=\fontfamily{pcr}\selectfont,%
  stringstyle=\tt,
  morestring=[b]",
  literate={\ }{$\mkern6mu$}1%
   {=}{{\tt\raisebox{-.15mm}{=}}}1%
   {[}{{\tt\raisebox{-.15mm}{[}}}1%
   {]}{{\tt\raisebox{-.15mm}{]}}}1%
   {->}{{$\rightarrow$}}1%
   {==>}{{$\Rightarrow$}}1%
   {wedge}{{$\wedge$}}1%
   {>=}{{$\geq$}}1%
   {'a}{{$\alpha$}}1%
   {'b}{{$\beta$}}1%
   {ldots}{$\ldots$}1%
   {!}{$\forall$}1%
   {``}{\hspace{-.5mm}`\hspace{-1mm}`}1%
  }

\lstdefinelanguage{SMLSmall}%
{columns=fullflexible,
  keywords={THEN,THENL,val,open,store_thm,fun,fn,let,in,end,true,
  while,do,if,then,else,break,return,;},%
  frame=none,
  sensitive=true,
  keywordstyle=\fontfamily{lmss}\scriptsize\selectfont,%
  basicstyle=\fontfamily{pcr}\small\selectfont,%
  stringstyle=\tt,
  morestring=[b]",
  literate={\ }{$\mkern6mu$}1%
   {=}{{\tt\raisebox{-.15mm}{=}}}1%
   {[}{{\tt\raisebox{-.15mm}{[}}}1%
   {]}{{\tt\raisebox{-.15mm}{]}}}1%
   {->}{{$\rightarrow$}}1%
   {==>}{{$\Rightarrow$}}1%
   {wedge}{{$\wedge$}}1%
   {<=}{{$\leq$}}1%
   {>=}{{$\geq$}}1%
   {'a}{{$\alpha$}}1%
   {'b}{{$\beta$}}1%
   {ldots}{$\ldots$}1%
   {emptyset}{$\emptyset$}1%
   {!}{$\forall$}1%
   {fff}{{DB.fetch}}1%
   {``}{\hspace{-.5mm}`\hspace{-1mm}`}1%
  }
\begin{document}
\maketitle

\begin{abstract}
We implement an automated tactical prover TacticToe on top of the HOL4 
interactive theorem
prover. TacticToe learns from human proofs which mathematical technique is
suitable in each proof situation.
This knowledge is then used in a Monte Carlo tree search algorithm to
explore promising tactic-level proof paths.
On a single CPU, with a time limit of 60 seconds, TacticToe proves 66.4\% of
the 7164 theorems in HOL4's standard library, whereas
E~prover with auto-schedule solves 34.5\%. The success rate rises to 69.0\% by
combining the results of TacticToe and E~prover.
\end{abstract}

\section{Introduction}
Many of the state-of-the-art interactive theorem provers (ITPs) such as
  \holfour~\cite{hol4}, \hollight~\cite{Harrison09hollight},
  \isabelle~\cite{isabelle}
  and \coq~\cite{coq-book} provide high-level parameterizable tactics for constructing proofs.
  Tactics analyze the current proof state (goal and
  assumptions) and apply non-trivial proof transformations.
  Formalized proofs take advantage of different levels of automation which are
  in increasing order of generality:
  specialized rules, theory-based strategies and general purpose strategies.
  Thanks to progress in proof
  automation, developers can delegate more and more complicated proof
  obligations to general purpose strategies. Those are implemented by automated
  theorem provers (ATPs) such as \eprover~\cite{eprover}. Communication
  between
  an ITP and ATPs is made possible by a ``hammer''
  system~\cite{hammers4qed,tgck-cpp15}. It acts as an interface by performing
  premise selection, translation and proof reconstruction.
  Yet, ATPs are not flawless and more precise user-guidance, achieved
  by applying a particular sequence of specialized rules, is almost always
  necessary to develop a mathematical theory.

  In this work, we develop in the ITP \holfour a procedure to select suitable
  tactics depending on the current proof state by learning
  from previous proofs. Backed by this machine-learned guidance, our prover
  \tactictoe executes a \emph{Monte
  Carlo tree search}~\cite{montecarlo} algorithm to find sequences of tactic
  applications
  leading to an ITP proof.

\subsection{The problem}
An ITP is a development environment for the construction of formal proofs in
which it is possible to write a proof as a sequence of
applications of primitive inference rules. This approach is not preferred
by developers because proving a valuable theorem requires often many
thousands of primitive inferences.
The preferred approach is to use and devise high-level tools and
automation that abstract over useful ways of producing proof pieces.
A specific, prevalent instance of this approach is the use of \emph{tactics}
for \emph{backward} or goal-directed proof.
Here, the ITP user operates on a proof state, initially the desired conjecture,
and applies tactics that transform the proof state until it is solved.
Each tactic performs a sound transformation of the proof state: essentially, it
promises that if the resulting proof state can be solved, then so can the
initial one. This gives rise to a machine learning problem: can we learn a
mapping from a given proof state to the next tactic (or sequence of tactics)
that will productively advance the proof towards a solution?

\paragraph{Goals and theorems.}
A goal (or proof state) is represented as a \emph{sequent}.
A sequent is composed of a set of assumptions and a conclusion, all of which
are higher-order logic formulas~\cite{gordon01hol}.
When a goal is proven, it becomes a theorem.
And when the developer has given the theorem a name, we refer to it as a
\textit{top-level} theorem.

\paragraph{Theories and script files.}
Formal developments in \holfour are organized into named theories, which are
implemented by \emph{script} files defining modules in the
functional programming language \sml.
Each script file corresponds to a single theory, and contains definitions of
types and constants as well as statements of theorems together with their
proofs.
In practice, tactic-based proofs are written in an informally specified
\emph{embedded domain-specific language}: the language consisting of
pre-defined tactics and \emph{tacticals} (functions that operate on tactics).
However, the full power of \sml is always available for users to define their
own tactics or combinations thereof.

\paragraph{Tactics in \holfour.}
A \emph{tactic} is a function
that takes a goal (or proof obligation) and returns a list of goals (subgoals
of the original goal) together
with a \textit{validation function}.
Calling the validation function on a list of \emph{theorems},
corresponding to
the list of subgoals, results in a theorem corresponding to the original goal.
For example, if the list of subgoals is empty, then calling the
validation function on the empty list yields the original goal as a theorem.
In this way, tactics implement the plumbing of goal-directed proof
construction, transforming the conjecture by reasoning in a backward manner.
Since validation functions are only important to check the final
proof, we omit them during the description of the proof search algorithm. We 
denote by $t(g)$ the list of goals produced by the application of a tactic $t$ 
to a goal $g$.

\subsection{Contributions}
This paper extends work described previously~\cite{tgckju-lpar17}, in which
we proposed the idea of a tactic-based prover based on supervised
learning guidance. We achieved a 39\% success rate on theorems of the
\holfour standard library by running \tactictoe with a 5 second timeout.
The contributions of this paper and their effect on our system are:

\begin{itemize}
\item Monte Carlo tree search (MCTS) replaces A* as our
  search algorithm (Section~\ref{sec:proofsearch}). The MCTS algorithm gives
more balanced feedback on
the success of each proof step. The policy and
value function are learned
through supervised learning.
\item Proof guidance required by MCTS is given by prediction algorithms
through supervised learning (Section~\ref{s:prediction}).
These predictors are implemented for three kinds of objects: tactics,
theorems, and lists of goals.
\item We introduce an orthogonalization process that eliminates redundant tactics (Section~\ref{sec:ortho}).
\item We introduce a tactic abstraction mechanism
  (Section~\ref{sec:synthesis}), which enables us to create
more general and flexible tactics by dynamically
predicting tactic arguments.
\item The internal ATP \metis is complemented
  by asynchronous calls to \eprover (Section~\ref{sec:atp}) to help \tactictoe
during proof search.
\item Proof recording at the tactic level is made more precise
  (Section~\ref{sec:recording}). We now support pattern
matching constructions and opening SML modules.
\item Evaluation of \tactictoe with a 60 seconds timeout achieves a 66\%
success rate on the standard library. Comparisons between
\tactictoe and \eprover on different type of problems are reported in
Section~\ref{s:experiments}.

\item Minimization and embellishment of the discovered proof facilitates user
  interaction (Section~\ref{sec:proofdisplay}).
\end{itemize}

\subsection{Plan}

\begin{figure}[h]
\begin{center}
\begin{tikzpicture}[node distance = 1.9cm]
  \node [] (aa) {\textbf{Learning}};
  \node [cloud, below of=aa, node distance = 1cm] (a) {Formal library};
  \node [block, below of=a] (b) {Proof recording};
  \node [cloud, below of=b] (c) {Knowledge base};
  \node [block, below of=c] (e) {Training loop};
  \node [block, below of=e] (d) {Tactic predictors};

  \draw [-to,black,thick] (a) to (b);
  \draw [-to,black,thick] (b.south) to (c.north);
  \draw [-to,black,thick] (c) to [bend left]  (e);
  \draw [-to,black,thick] (e) to [bend left]  (c);
  \draw [-to,black,thick] (c.east) to [out=330,in=30] (d);

  \node [right of=aa, node distance=6.2cm] (0) {\textbf{Proving}};
  \node [cloud, below of=0, node distance = 1cm] (1) {Conjecture};
  \node [cloud, below of=1, node distance = 1.7cm] (2) {Search tree};
  \node [block, below of=2, node distance=1.3cm] (3) {Tactic policy};
  \node [above of=2, xshift=-16mm, node distance = 0.9cm] (00) {MCTS proof
  search};
  \node [block, below of=3, node distance = 1.3cm] (4) {Tactic evaluation};
  \node [block, fit=(2) (3) (4)] (5) {};
  \node [block, below of=4, node distance = 1.7cm] (6) {Proof minimization};
  \node [cloud, below of=6, node distance = 1.6cm] (7) {Proof};
  \draw [-to,black,thick] (1) to (5);
  \draw [-to,black,thick] (5) to (6);
  \draw [-to,black,thick] (6) to (7);
  \draw [-to,black,thick] (d) to [out=0, in=180] (3);
  \draw [-to,black,thick] (d) to [out=0, in=180] (4);

\end{tikzpicture}
\end{center}
\caption{\label{fig:components}Relation between modules of the learning and
proving toolchains. Functions are represented by rectangles and objects by
ellipses.}
\end{figure}
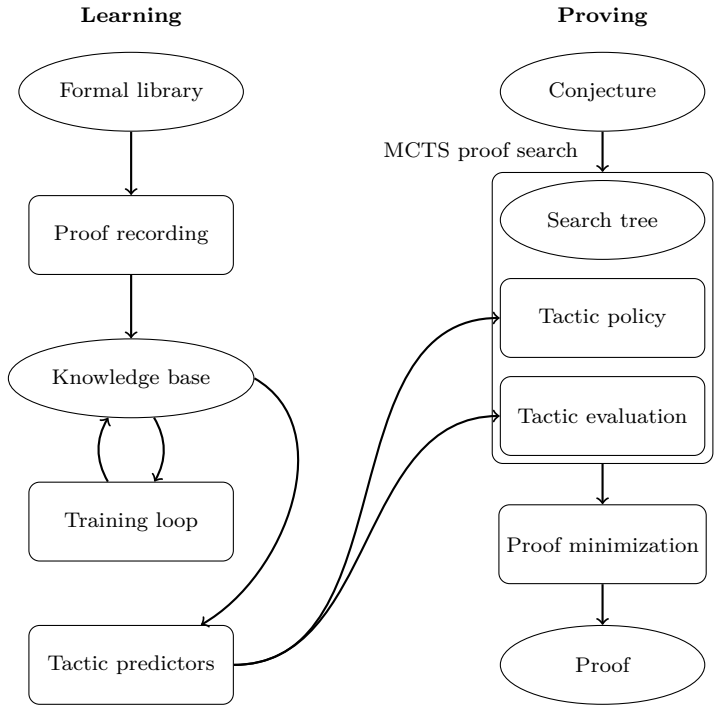

The different components depicted in Figure~\ref{fig:components} give a
high-level overview of TacticToe and our approach to generating tactic-level proofs by learning from recorded knowledge.
On one side, we have the learning aspect of TacticToe whose purpose is to produce a high quality function to predict a tactic given a goal state.
Tactic prediction uses a knowledge base that is gleaned, in the first place, from human-written proofs.
On the other side, we have the proving aspect of TacticToe, which uses tactic prediction to guide a proof search algorithm in the context of a given conjecture.

This paper is organized firstly around the proving aspect:
We define the proof search tree (Section~\ref{sec:prelim}), explain the essence of our approach to learning to predict tactics (Section~\ref{s:prediction}), and then present the prediction-guided proof search algorithm (Section~\ref{sec:proofsearch}).
We also describe our preselection method (Section~\ref{sec:presel}) for speeding up prediction during search.
Afterwards, we delve into some details of the learning aspect of TacticToe:
We describe approaches to improving the knowledge base that supports prediction (orthogonalization in Section~\ref{sec:ortho} and abstraction in Section~\ref{sec:synthesis}), and describe how we build the knowledge base in the first place by extracting and recording tactic invocations from a library of human-written proofs
(Section~\ref{sec:recording}).

In Section~\ref{s:experiments}, we present an evaluation of TacticToe on a large set of theorems originating from \holfour.
Finally, we present a feature of TacticToe that makes it more suitable for use by human provers, namely, minimization and prettification (Section~\ref{sec:proofdisplay}) and compare TacticToe's generated proofs to human proofs (Section~\ref{sec:case_study}).

\section{Search Tree}\label{sec:prelim}
The application of a tactic to an initial conjecture produces a list of goals.
Each of the created goals can in turn be the input to other tactics.
Moreover, it is possible to try multiple tactics on the same goal.
In order to keep track of progress made from the initial conjecture, we organize goals and tactics into a graph with lists of goals
as nodes and tactics as edges (See Definition~\ref{def:stree}).

In this section, we only give a description of the graph at a given
moment of the search, after some number of tactic applications.
Construction of the tree is done by the MCTS algorithm in
Section~\ref{sec:proofsearch}. The MCTS algorithm is guided by prediction
algorithms presented in Section~\ref{s:prediction}.

\begin{figure}{}
\begin{center}
\begin{tikzpicture}[node distance=0.7cm]
\node [tcircle] (0) {$g_i$};
\node [right of=0] (0r) {...};
\node [tcircle,right of=0r] (0rr) {$g_n$};
\node [left of=0] (0l) {...};
\node [tcircle,left of=0l] (0ll) {$g_0$};
\node [draw,fit=(0rr) (0ll)] (0b) {};
\node [left of=0ll] {$a_0$};
\node [above of=0,node distance=2cm] (2) {...};
\node [left of=2, tcircle] (2l) {$\phantom{g_3}$};
\node [tcircle,right of=2] (2r) {$\phantom{g_3}$};
\node [draw, fit=(2l) (2r)] (2b) {};
\node [left of=2l] {$a_j$};

\node [tcircle, left of=2l, node distance=3cm] (1r) {$\phantom{g_3}$};
\node [left of=1r] (1) {...};
\node [tcircle, left of=1] (1l) {$\phantom{g_3}$};
\node [draw, fit=(1l) (1r)] (1b) {};
\node [left of=1l] {$a_1$};

\node [right of=2r,node distance=3cm] (3l) {$\phantom{g_3}$};
\node [right of=3l] (3) {\phantom{...}};
\node [right of=3] (3r) {$\phantom{g_3}$};
\node [fit=(3l) (3r)] (3b) {};

\node [fit=(1r) (2l)] (1m) {};
\node [fit=(2l) (3r)] (2m) {};

\draw[-to,thick] (0) to node[xshift=-10](t1){$t_1$} (1b);
\draw[-to,thick] (0) to node[xshift=-4](t2){$t_j$} (2b);
\draw[-to,dotted,thick] (0) to (1m);
\draw[-to,dotted,thick] (0) to (2m);
\draw[-to,dotted,thick] (0) to node[xshift=15](t2){$t_m$} (3b);
\end{tikzpicture}
\end{center}
\caption{\label{fig:choice}A node of a search tree $a_0$, the list of goals
$g_0 \ldots g_n$ it
contains and the nodes $a_1 \ldots a_m$ derived from the application of the
tactics $t_1 \ldots t_m$ to $g_i$.}
\end{figure}
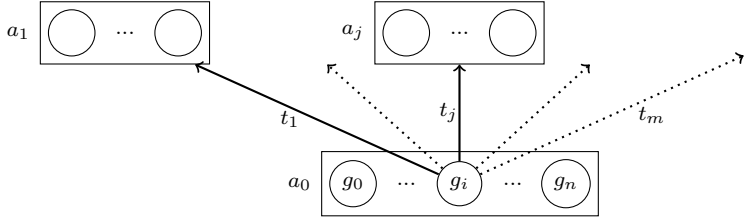

\begin{definition}\label{def:stree}(search tree)\\
A search tree $\mathfrak{T}$ is a sextuple
$(\mathbb{T},\mathbb{G},\mathbb{A},T,G,A)$
that respects the following conditions:
\begin{itemize}
\item $\mathbb{T}$ is a set of tactics, $\mathbb{G}$ is a set of goals
 and $\mathbb{A}$ is a set of nodes representing lists of goals. All objects
 are tagged with their position in the tree (see Figure~\ref{fig:choice}).
\item $T$ is a function from $\mathbb{G}$ to $\mathcal{P}(\mathbb{T})$. It
takes a goal and return the set of tactics already applied to this goal.
\item $G$ is a function from $\mathbb{A}$ to $\mathcal{P}(\mathbb{G})$.
It takes a node and returns the list of goals of this node.
\item $A$ is a function that takes a pair $(g,t)\in \mathbb{G} \times
\mathbb{T}$ such as $t \in T(g)$ and
returns a node $A(g,t)$ such that $t(g) = G(A(g,t))$. In other words, the output
$t(g)$ is exactly the list of goals contained in $A(g,t)$.
\item It is acyclic, i.e., a node cannot be a strict descendant (see
Definition~\ref{def:desc}) of itself.
\item There is exactly one node with no parents. This root node contains
exactly one goal which is the initial goal (conjecture).
\end{itemize}

\end{definition}

If no explicit order is given, we assume that the sets
$\mathbb{T},\mathbb{G},\mathbb{A}$ are equipped with an
arbitrary \textit{total order}. Figure~\ref{fig:choice} depicts part of a
search tree. Goals, nodes and tactics are represented
respectively by circles, rectangles and arrows.

In the following, properties are defined in the context of a
search tree $\mathfrak{T}$. In Definition~\ref{def:solved}, we specify
different states for a goal: open, pending or solved. There, we define
what a solved goal and solved node are by mutual recursion on the number of
steps it takes to solve a goal and a node.
\begin{definition}\label{def:solved}(solved goal, solved nodes, open goal,
pending goal)\\
The set of \textit{solved nodes} $\mathbb{A}^*$ and
the set of \textit{solved goals} $\mathbb{G}^*$ are defined inductively by:

\begin{align*}
\mathbb{A}^{0} &=_{\mathrm{def}}
\lbrace a \in \mathbb{A}\ |\ G(a) = \emptyset \rbrace \\
\mathbb{G}^{0} &=_{\mathrm{def}} \lbrace g \in \mathbb{G}\ |\
\exists t \in T(g).\ A(g,t) \in \mathbb{A}^{0} \rbrace\\
\mathbb{A}^{n+1} &=_{\mathrm{def}} \lbrace a \in \mathbb{A}\ |\
\forall g \in G(a).\ g \in \mathbb{G}^{n} \rbrace\\
\mathbb{G}^{n+1} &=_{\mathrm{def}} \lbrace g \in \mathbb{G}\ |\
\exists t \in T(g).\ A(g,t) \in \mathbb{A}^{n+1} \rbrace \\
\mathbb{A}^* &=_{\mathrm{def}} \bigcup_{n \in \mathbb{N}} \mathbb{A}^n \ \ \ \ \
\mathbb{G}^* =_{\mathrm{def}} \bigcup_{n \in \mathbb{N}} \mathbb{G}^n\\
\end{align*}

We call a goal \textit{unsolved} if it is not in the set
of solved goals. Similarly, we call a node \textit{unsolved} if it does not
belong to the set of solved nodes.
The \textit{open goal} of a node is the first unsolved goal of this node
according to
some preset order. The other unsolved goals are called \textit{pending goals}.
If a node is unsolved, then it contains a unique open goal.
During MCTS exploration, we do not explore pending goals of an unsolved node
before its open goal is solved which justifies our terminology. In other words,
our search tree has the following property: if $g$ is a pending goal then
$T(g)$ is empty.

\end{definition}

In Definition~\ref{def:desc}, we define relations between goals
and nodes in terms of parenthood.

\begin{definition}\label{def:desc}(children, descendant, ancestors)\\
The \textit{children of a goal} $g$ is a set of nodes defined by:
\[\mathit{Children}(g) = \lbrace A(g,t)\in \mathbb{A}\ |\ t \in T(g) \rbrace\]

By extension, we define the \textit{children of a node} $a$ by:
\[\mathit{Children}(a) = \bigcup_{g \in G(a)} \mathit{Children}(g) \]

The \textit{descendants of a node} $a$ by:
\begin{align*}
\mathit{Descendants}^{0}(a) &=_{\mathrm{def}} \lbrace a \rbrace \\
\mathit{Descendants}^{n+1}(a) &=_{\mathrm{def}} \bigcup_{a' \in \mathit{Descendants}^{n}(a)}
\mathit{Children}(a') \\
\mathit{Descendants}(a) &=_{\mathrm{def}} \bigcup_{n \in \mathbb{N}} \mathit{Descendants}^n(a)\\
\end{align*}

And the \textit{ancestors of a node} $a$ by:
\[\mathit{Ancestors}(a) =_{\mathrm{def}} \lbrace b \in \mathbb{A} \ | \ a \in
Descendants(b) \rbrace\]
\end{definition}

In Definition~\ref{def:prod}, we decide if a tactic is productive based on its
contribution to the search tree.

\begin{definition}\label{def:prod} (productive tactic)
The application of a tactic $t$ on an open goal $g$ is called
\textit{productive} if and only if
all the following conditions are satisfied:
\begin{itemize}
\item It does not fail or timeout. To prevent tactics from looping, we
interrupt tactics after a short amount of time (0.05 seconds).
\item It does not loop, i.e., its output does not contain any goals that appear
in the ancestors of the node of the current goal.
\item It is not a redundant step, i.e., there does not exist $t'$ in $T(g)$
such as $t'(g) \subseteq t(g)$.
\end{itemize}

The third point of the definition is a partial attempt at preventing confluent
branches. The general case where two branches join after $n$ steps is handled
by a tactic cache which memorizes tactic effects on goals. This
cache allows faster re-exploration of confluent branches which remain
separated in the search tree.
\end{definition}

\section{Prediction}\label{s:prediction}
The learning-based selection of relevant lemmas significantly improves the
automation in hammer systems~\cite{BlanchetteGKKU16}. In \tactictoe we use the
distance-weighted \emph{$k$ nearest-neighbour} classifier~\cite{DudaniS76}
adapted for lemma selection~\cite{ckju-pxtp13}. It allows for hundreds of
predictions per second, while maintaining very high prediction quality~\cite{femalecop}.
We rely on this supervised machine learning algorithm to predict objects of
three kinds: tactics, theorems and lists of goals. The number of predicted
objects $k$ is
 called the \emph{prediction radius}. For a goal $g$, we denote
by $\mathit{Predictions}^{\mathit{tactic}}_k (g)$ (respectively
$\mathit{Predictions}^{\mathit{theorem}}_k (g)$
and $\mathit{Predictions}^{\mathit{goal\_list}}_k (g)$) the $k$ tactics
(respectively theorems and lists of goals) selected by our prediction algorithm.

We discuss below the specifics associated with the prediction of each kind of object.
In particular, we present the dataset from which objects are selected and the
purpose of the selection process. The similarity measure backing the
predictions is described in Section~\ref{sec:features} and three methods for
improving the speed and quality of the predictions are presented in
Sections~\ref{sec:presel}, \ref{sec:ortho} and \ref{sec:synthesis}.

\paragraph{Tactics}
To start with, we build a database of tactics
$\mathbb{D}_{\mathit{tactic}}$ consisting
of tactic-goal pairs recorded from successful tactic applications in human
proofs (recording will be discussed in
Section~\ref{sec:recording}).
During proof search, the recorded goals will be reordered according to their
similarity with a target goal $g$ (usually the open goal of a node).
The recorded pairs and their similarity to $g$ induce an
order on tactics. Intuitively, tactics which have been successful on
goals similar to $g$ should be tried first. Indeed, they are more likely to lead
to a proof.
This predicted tactic order is then transformed into a prior policy of our MCTS
algorithm (see Section~\ref{sec:policy}).
Tactic selection is also used to improve the quality of the database
of tactics during orthogonalization (see Section~\ref{sec:ortho}).

\paragraph{Theorems as Arguments of Tactics}
We first collect a database of theorems $\mathbb{D}_{\mathit{theorem}}$ for our 
predictor to select from.
It includes the \holfour theorem database and theorems from the local namespace.
Predicted tactics during proof search and orthogonalization may include
tactics where arguments (lists of theorems) have been erased (see Section
\ref{sec:synthesis}).
We instantiate those tactics by theorems that are syntactically the closest to
the goal as they are more likely to help.
The same algorithm selects suitable premises for ATPs integrated with
\tactictoe (see Section~\ref{sec:atp}).

\paragraph{Lists of Goals as Output of Tactics}
We compile a dataset of tactic outputs $\mathbb{D}_{\mathit{goallist}}$ during 
orthogonalization (see
Section~\ref{sec:ortho}).
Some elements of this set will be considered positive examples (see
Section~\ref{sec:evaluation}).
During proof search, given a list of
goals $l$ created by a tactic, we select a set of lists of goals that are most
similar to $l$. The ratio of positive examples in the selection gives us a
prior evaluation for MCTS.

\subsection{Feature Extraction and Similarity Measures}\label{sec:features}

We predict tactics through their associated goals. So, the features
for each kind of object can be extracted from different representations of
mathematical formulas. We start by describing features for \holfour terms. From
there, we extend the extraction mechanism
to goals, theorems and lists of goals. Duplicated features are always removed
so that each object has an associated set of features.
Here are the kinds of features we extract from terms:
\begin{itemize}
\item names of constants, including the logical operators,
\item type constructors present in the types of constants and variables,
\item subterms with all variables replaced by a single placeholder $V$,
\item names of the variables.
\end{itemize}
Goals (respectively theorems) are represented by pairs consisting of a list of
terms (assumptions) and a term (conclusion). We distinguish between
features of the assumptions and features of the conclusion by adding a
different tag to elements of each set. The feature of a goal (respectively
theorem)
are the union of all these tagged features. From
that, we can construct features for a list of goals by computing the union of
the features of each goal in the list.

We estimate the similarity between two objects $o_1$ and $o_2$
through their respective feature sets $f_1$ and $f_2$.
The estimation is based on the frequency of the features in the intersection of
$f_1$ and $f_2$. A good rule of thumb is that the rarer the shared features
are, the more similar two goals should be. The relative rarity of each feature
can be estimated by calculating its TF-IDF weight~\cite{Jones72astatistical}.
We additionally raise these weights to the sixth power giving
even more importance
to rare features~\cite{BlanchetteGKKU16}.

The first similarity measure $\mathit{sim}_1$ sums the weight of all shared
features to compute the total score.
In a second similarity measure $\mathit{sim}_2$, we additionally take
the total number of features into account, to reduce the seemingly unfair
advantage of big
feature sets in the first scoring function.
\[\mathit{sim}_1 (o_1, o_2) = {\sum\nolimits_{\,f \in f_0 \cap
f_1}{\mathrm{tf\_idf}(f)^{6}}}\]
\[\mathit{sim}_2 (o_1, o_2) = \frac{{\sum\nolimits_{\,f \in f_0 \cap
f_1}{\mathrm{tf\_idf}(f)^{6}}}}
{ln (e + |f_0| + |f_1|)}\]

In our setting, making predictions for an object $o$ consists of sorting a set
of objects by their similarity to $o$. We use $\mathit{sim}_1$ for tactics and theorems
and $\mathit{sim}_2$ for lists of goals. The reason why $\mathit{sim}_1$ is used for
predicting tactics is that tactics effective on a large goal $g$ are often also
suitable for a goal made of sub-formulas of $g$. The same monotonicity
heuristic can justify the use of $\mathit{sim}_1$ for theorems. Indeed, in practice a
large theorem is often a conjunction of theorems from the same domain.
And if a conjunction contains a formula related to a problem, then the other
conjuncts from the same domain may also help to solve it.

\section{Proof Search}\label{sec:proofsearch}

Our prediction algorithms are not always accurate.
Therefore, a selected tactic may
fail, proceed in the wrong direction or even loop.
For that reason, predictions need to be accompanied by a
proof search mechanism that allows for backtracking and
can choose which proof tree to extend next and in which direction.
Our search algorithm is a Monte Carlo tree search~\cite{montecarlo} algorithm
that relies on a prior evaluation function and a prior policy function. Those 
priors are learned through direct supervised learning for the policy and via 
the data accumulated during orthogonalization for the evaluation. The
prior policy and evaluation are estimated by the simple supervised k-NN 
algorithm. We call them priors because these function are known a priori before 
starting the proof search. In contrast, we define later in this section a 
current policy and a current evaluation which are continuously updated by the 
MCTS algorithm during proof search through a feedback mechanism. A 
characteristic of the MCTS 
algorithm is that it offers a good 
trade-off between exploitation and exploration. This means that the algorithm 
searches deeper the more promising branches and leaves enough time for the 
exploration of less likely alternatives.

\begin{remark}
Neural networks trained through reinforcement learning can be
very effective for approximating the policy and evaluation as demonstrated in
.e.g. AlphaGo Zero~\cite{silver2017mastering}.
But training neural networks is
computationally expensive and has not yet been
proven effective in our context. That is why we chose a simpler machine
learning model for our project.
\end{remark}

We first describe the proof search algorithm and explain later how to compute
the prior policy \texttt{PriorPolicy} and prior evaluation
\texttt{PriorEvaluation} functions.

\subsection{Proof exploration}

The proof search starts by creating a search tree during initialization.
The search tree then evolves by the repetitive applications of MCTS steps. A
step in the main loop of MCTS consists of three parts: node selection, node
extension, and backpropagation. And the decision to stop the loop is taken
by the resolution module.

\paragraph{Initialization}
The input of the algorithm is a goal $g$ (also called conjecture) that we
want to prove. Therefore the search tree starts with only one node containing
the list of goals
$[g]$.

\paragraph{Node selection}
Through node extension steps the search tree grows and the number of paths to
explore increases. To decide which node to extend next, the MCTS algorithm
computes for each node a value (see Definition~\ref{def:value}) that changes
after each MCTS step.
The algorithm that performs
node selection starts from the root of the search tree. From the current node,
the function \texttt{CompareChildren} chooses the child node with the highest
value among the children of its open goal.

If the highest children value is higher than the \textit{widening policy} (see
Section~\ref{sec:policy}), then the
selected child
becomes the current node, otherwise the final result of node selection is the
current node. The following pseudo-code illustrates our node selection
algorithm:

\begin{lstlisting}[language=SMLSmall]
CurrentNode = Root(Tree);
while true do
  if Children(CurrentNode) = emptyset then break;
  (BestChild, BestValue) = CompareChildren (CurrentNode);
  if WideningPolicy (ChosenNode) >= BestValue then break;
  CurrentNode = BestChild
end;
return CurrentNode;
\end{lstlisting}

\begin{definition}\label{def:value}(Value)\\
The value of the $i^{th}$ child $a_i$ of the open goal of a parent node $p$
is determined by:
  \[\mathit{Value}(a_i) = \mathit{CurEvaluation}(a_i) + c_{\mathit{exploration}} *
\mathit{Exploration}(a_i)\]

The current evaluation $CurEvaluation(a_i)$ is the average evaluation of
all descendants of $a_i$ including node extension failures.

\[\mathit{CurEvaluation}(a_i) =
  \sum_{a' \in \mathit{Descendants}(a_i)} \frac{\mathit{PriorEvaluation}(a')}
  {|\mathit{Descendants}(a_i)| + \mathit{Failure}(a_i)}\]

where the number $\mathit{Failure}(a_i)$ is the number of failures that occurred
during node extension from descendants of $a_i$ .

The exploration term is determined by the prior policy and the current policy.
The current policy is calculated from the number of times a node $x$ has been
traversed during node selection, denoted $\mathit{Visit}(x)$.

\[\mathit{Exploration}(a_i) =
\frac{\mathit{PriorPolicy}(a_i)}{\mathit{CurPolicy}(a_i)}\]

\[\mathit{CurPolicy}(a_i) = \frac{1 +
\mathit{Visit}(a_i)}{\sqrt{\mathit{Visit}(p)}}\]
The policy can be seen as a skewed percentage of the number of times a node was
visited. The square root favors exploration of nodes with
  few visits. The coefficient $c_{\mathit{exploration}}$ is experimentally determined and
adjusts the trade-off between exploitation and exploration.
\end{definition}

\paragraph{Node extension}
Let $a$ be the result of node selection.
If $a$ is a solved node or the descendant of a solved node, then extending $a$
would not be productive and the algorithm reports a failure for this MCTS step.
If $a$ is not solved, it applies the best untested tactic $t$ on the open
goal $g$ of this
node according to the prediction order for $g$.
If no such tactic exists or if $t$ is not productive, the algorithm returns a
failure.
If node extension succeeds, a new node containing $t(g)$ is added to the
children of $a$.

\paragraph{Backpropagation}
During backpropagation we update the statistics of all the nodes traversed or
created during this MCTS step:
\begin{itemize}
\item Their visit number is incremented.
\item If node extension failed, their failure count is incremented.
\item If node extension succeeded, they inherit the evaluation of the created
child.
\end{itemize}

These changes update the current evaluation and the current policy of the
traversed nodes.
After completing backpropagation, the process loops back to node
selection.

\paragraph{Resolution}
The search ends when the algorithm reaches one of these 3 conditions:
\begin{itemize}
\item It finds a proof, i.e., the root node is solved. In this case,
the search returns a minimized and
prettified tactic proof (see Section~\ref{sec:proofdisplay}).
\item It saturates, i.e., there are no tactics to be applied to any open goal.
This occurs less than 0.1 percent of the time in the full-scale experiment.
And this happens only at the beginning of the \holfour library, where the
training
data
has very few tactics.
\item It exceeds a given time limit (60 seconds by default).
\end{itemize}

\subsection{Supervised learning guidance}
Our MCTS algorithm relies on the guidance of two supervised learning methods.
The prior policy estimates the probability of success of a tactic $t$ on a
goal $g$ and associates it with the potential children that would be created by
the
application of $t$ to $g$. The prior evaluation judges if a branch is fruitful
by analyzing tactic
outputs. Both priors influence the rates at which branches of the search tree
are explored.

\paragraph{Prior policy}\label{sec:policy}
Let $\mathfrak{T}$ be a search tree after a number of MCTS steps.
Let $n$ be the number of predictions returned by the prior policy (500 in our 
experiments).
Let $p$ be a selected parent node in $\mathfrak{T}$ and $g$ its open goal. We 
order the list of $n$ productive tactics already
applied to $g$ by their similarity score. Indeed, a tactic is likely to 
be useful to solve $g$ if it has been applied on a goal similar to 
$g$ in a previous proofs. 
We note the resulting list
$t_0,\ldots,t_{n-1}$.
Let $c_{\mathit{policy}}$ be a constant heuristic optimized during our experiments.
We calculate the prior policy of a child $a_i$ produced by the tactic
$t_i$ by:
\[\mathit{PriorPolicy}(a_i) = (1 - c_{\mathit{policy}})^{i} * c_{\mathit{policy}}\]

In order to include the possibly of trying more tactics on $g$ we define the
widening policy on the parent $p$ for its open goal $g$ to be:

\[\mathit{WideningPolicy}(p) = (1 - c_{\mathit{policy}})^{n} * 
c_{\mathit{policy}}\]

\paragraph{Prior evaluation}\label{sec:evaluation}

We now concentrate on the definition of a reasonable evaluation function for the
output of tactics. Intuitively, a list of goals is worth exploring further if
we believe there exists a short proof for it. We could estimate the likelihood
of finding such a proof from
previous proof searches. However, extracting lists of goals from all proof
attempts creates too much data which slows down the prediction algorithm.
So, we only
collect outputs of tactics tested during orthogonalization (see
Section~\ref{sec:ortho}). We
declare a list of goals $l$ to be \textit{positive} if it has been produced
by the winner of an orthogonalization competition.
The set of positive examples in
$\mathit{Predictions}^{\mathit{goal\_list}}_k(l)$ is denoted
$\mathit{Positive}^{\mathit{goal\_list}}_k(l)$.
We chose $k=10$ as a default evaluation radius in our experiments.
And we evaluate a node $a$ through the list of goals $G(a)$ it
contains using the prior evaluation function:
\begin{align*}
\mathit{PriorEvaluation}_k (a) &=
  \frac{|\mathit{Positive}^{\mathit{goal\_list}}_k(G(a))|}{k}\\
\end{align*}

\subsection{ATP Integration}\label{sec:atp}
General-purpose proof automation mechanisms which combine proof translation to
ATPs with machine learning (``hammers'') have become quite successful in
enhancing the automation level in proof assistants~\cite{hammers4qed}.
Since external automated reasoning techniques sometimes outperform the combined
power of tactics, we combine the \tactictoe search with
general purpose provers such as the ones found in \holyhammer for
\holfour~\cite{tgck-cpp15}.

\holfour already integrates a superposition-based prover \metis~\cite{metis}
for this purpose. It is already
recognized by the tactic selection mechanism in \tactictoe and thanks to
abstraction, its
premises can be predicted dynamically. Nevertheless,
we think that the performance of \tactictoe can be boosted by giving the ATP
\metis a special status among tactics. This arrangement consists of always
predicting
premises for \metis, giving it a slightly higher timeout and
trying it first on each open goal. Since \metis does not create
new goals, these modifications only induce an overhead that increases linearly
with the number of nodes.

In the following, we present the implementation of asynchronous calls to
\eprover~\cite{eprover} during \tactictoe's proof search.
Other external ATPs can be integrated with \tactictoe in a similar manner.

First, we develop a tactic which expects a list of premises as an
argument and calls \eprover on the goal together with the premises translated
to a first-order problem. If \eprover
succeeds on the problem, the premises used in the external proof are used to
reconstruct the proof inside \holfour with \metis. Giving a special status
to \eprover is essential as \eprover calls do not appear in human-written
tactic proofs. Here, we use an even higher timeout (5 seconds) and
a larger number of predicted premises (128). We also try \eprover
first every time a new goal is created. Since calls to \eprover are
computationally expensive, they are run in parallel and asynchronously. To
update the MCTS values, the
result of each thread is back-propagated in the tree after completion. This
avoids slowing down \tactictoe's search loop. The number of asynchronous calls
to \eprover that can be executed at the same time is limited by the
number of cores available to the user.

\section{Preselection}\label{sec:presel}

In order to speed up the predictions during the search for a proof of a
conjecture $c$, we preselect 500 tactic-goal pairs and 500 theorems and a
larger number of lists of goals induced by the selection of tactic-goal pairs.
Preselected objects are the only objects available to \tactictoe's search
algorithm for proving $c$.

The first idea is to select tactic-goal pairs and theorems by their similarity
with $c$ (i.e. $\mathit{Predictions}^{\mathit{tactic}}_{500} (c)$ and
$\mathit{Predictions}^{\mathit{theorem}}_{500} (c)$).
However, this selection may not be adapted at later stages of
the proof where goals may have drifted significantly from $c$. In order to
anticipate the production of diverse goals, our prediction during preselection
takes dependencies between objects of the same dataset into account.
This dependency relation is asymmetric. Through the relation, each object has
multiple children and at most one parent.
Once a dependency relation is established (see Definition~\ref{def:tacdep} and 
Definition~\ref{def:thmdep}), we can calculate a dependency score
for each object, which is the maximum of its similarity score
and the similarity score of its parent. Finally, the 500 entries with
highest dependency score are preselected in each dataset.

\begin{definition}\label{def:tacdep}(Dependencies of a tactic-goal pair)\\
Let $\mathbb{D}_{\mathit{tactic}}$ be the database of tactic-goal pairs.
The dependencies $D^*$ of a tactic-goal pair $(t_0,g_0)$ are
inductively defined by:

\begin{align*}
D_0 &=_{\mathrm{def}} \lbrace (t_0,g_0) \rbrace \\
D_{n+1} &=_{\mathrm{def}} D_n \cup \lbrace (t,g)\in 
\mathbb{D}_{\mathit{tactic}}\  |\ \exists
(t',g') \in D_n.\ g \in t'(g') \rbrace  \\
D^* &=_{\mathrm{def}} \bigcup_{i \in \mathbb{N}} D_i\\
\end{align*}
\end{definition}

\begin{definition}\label{def:thmdep}(Dependencies of a theorem)\\
The dependencies of a theorem $t$ are the set of top-level theorems in
the database of theorems $\mathbb{D}_{\mathit{theorem}}$ appearing in the proof 
of $t$. These 
dependencies are recorded by tracking the use of top-level theorems through the 
inferences of the
kernel~\cite{tgck-cpp15}.
\end{definition}

We do preselection also for lists of goals.
To preselect a list of goals $l$ from our database of lists of goals 
$\mathbb{D}_{\mathit{goallist}}$, we
consider the tactic input $g$ that was at the origin of the production of $l$
during orthogonalization. The list $l$ is preselected if and only if $g$
appears in the 500 preselected tactic-goal pairs.

\section{Orthogonalization}\label{sec:ortho}
Different tactics may transform a single goal in the same way. Exploring such
equivalent paths
is undesirable, as it leads to inefficiency in automated proof search.
To solve this problem, we modify the construction of the tactics database.
Each time a new tactic-goal pair $(t,g)$ is extracted from a tactic proof and
about to be recorded, we consider if there does not already exist a better
tactic for $g$ in our database. To this end, we organize a competition between
the
$k$ closest tactic-goal pairs $\mathit{Predictions}^{\mathit{tactic}}_k(g)$.
In our experiments,
the default orthogonalization radius is $k=20$.
The winner is the tactic that subsumes (see Definition~\ref{def:tacsub}) the
original tactic on $g$ and that
appears in the largest number of tactic-goal pairs in the database.
The winning tactic $w$ is then associated with $g$, producing the pair $(w,g)$
and is stored in the database instead of the original pair $(t,g)$.
As a result, already successful tactics with a large coverage are preferred,
and new tactics are considered only if they provide a different contribution.
We now give a formal definition of the concepts of subsumption and coverage
that are required for expressing the orthogonalization algorithm.

\begin{definition} (Coverage)\\
Let $\mathfrak{T}$ be the database of tactic-goal pairs. We define the
coverage $\textit{\mbox{Coverage}}(t)$ of a tactic $t$ by the number of times
this tactic
appears in
$\mathfrak{T}$. Expressing this in a formula, we get:
\[\textit{\mbox{Coverage}}(t) =_{\mathrm{def}} |\lbrace g\ |\ (t,g) \in
  \mathfrak{T} \rbrace| \]
Intuitively, this notion estimates how general a tactic is by counting the
number of different goals it is useful for.
\end{definition}

\begin{definition} (Goal subsumption)\\
A goal subsumption $\le$ is a partial relation on goals.
Assuming we have a way to estimate the number of steps required to solve a goal,
a 
general and useful subsumption definition is for example:
\[g_1 \le g_2  \Leftrightarrow_{\mathrm{def}} g_1 \mbox{ has a proof with a length
shorter
or equal to the proof of }
g_2\]
In this example, the definition of the 
length of a proof is not specified, different definitions may produce different 
subsumptions.
By default, we however choose for efficiency reasons a minimal subsumption defined by:
\[g_1 \le g_2  \Leftrightarrow_{\mathrm{def}} g_1 \mbox{ is }\alpha\mbox{-equivalent
to } g_2\]
\end{definition}

We can naturally extend any goal subsumption to a subsumption on
lists of goals.
\begin{definition} (Goal list subsumption)\\
A goal list subsumption is a partial relation on lists of goals.
Given two lists of goals $l_1$ and $l_2$, we define it from the goal
subsumption $\le$ by:
\[l_1 \le l_2  \Leftrightarrow_{\mathrm{def}} \forall g_1 \in l_1.\ \exists g_2 \in l_2.\
g_1 \le g_2\]
\end{definition}

This allows us to define subsumption for tactics.
\begin{definition}\label{def:tacsub}(Tactic subsumption)\\
Given two non-failing tactics $t_1$ and $t_2$ on $g$, a tactic $t_1$ subsumes a
tactic $t_2$ on a goal $g$, denoted $\le_g$, when:

\[t_1 \le_g t_2 \Leftrightarrow_{\mathrm{def}} t_1(g) \le t_2(g)\]

If one of the tactics is failing then $t_1$ and $t_2$ are not comparable through
this relation.
\end{definition}

Finally, the winning tactic of the orthogonalization competition can be
expressed by the formula:

\[\mathbb{O}(t,g) = \argmax_{x\ \in\
\mathit{Predictions}^{\mathit{tactic}}_k(g) \cup
\lbrace t
\rbrace} \lbrace
\textit{\mbox{Coverage}}(x)\
|\ x \le_g t\rbrace\]

A database built with the orthogonalization process thus contains
$(\mathbb{O}(t,g),g)$ instead of $(t,g)$.

\section{Abstraction}\label{sec:synthesis}
One of the major weaknesses of the previous version of 
\tactictoe~\cite{tgckju-lpar17} was that
it could not create its own tactics. Indeed, sometimes no previous tactic is
adapted for a certain goal and creating a new tactic is necessary.
In this section we present a way to create tactics with different arguments
by abstracting them and re-instantiating them using a predictor for that kind
of argument. In the spirit of the orthogonalization method, we will try to
create tactics that are more general but have the same effect as the
original one. The generalized tactics are typically slower than their
original variants, but the added flexibility is worthwhile in practice. Moreover,
since we impose a timeout on each tactic (0.05 seconds by default), very slow
tactics fail and thus are not selected.

\subsection{Abstraction of Tactic Arguments}
The first step is to abstract arguments of tactics by replacing them by a
placeholder, creating a tactic with an unspecified argument.
Suppose we have a recorded tactic $t$. Since $t$ is a \sml code tree, we can
try to abstract any of the
\sml subterms. Let $u$ be a \sml subterm of $t$, and $h$ a variable (playing
the role of a placeholder), we
can replace $u$ by $h$ to create an abstracted tactic.
We denote this intermediate tactic $t[h/u]$. By repeating the
process, multiple
arguments may be abstracted in the same tactic. Ultimately, the more
abstractions are performed the more general a tactic is, as many tactics
become instances of the same abstracted tactic. As a consequence, it becomes
harder and harder to predict suitable arguments.

\subsection{Instantiation of an Abstracted Tactic}
An abstracted tactic is not immediately applicable to a goal, since it contains
unspecified arguments. To apply an abstracted tactic $\hat{t}$, we first need
to instantiate the placeholders $h$ inside $\hat{t}$. Because it is difficult to
build a general predictor effective on each type of argument, we manually
design different argument predictors for each type. Those predictors are
given a goal $g$ as input and return the best predicted argument for this
goal. We discuss here specific algorithm for instantiating three common types 
of arguments: theorem list, theorem, term. Creating predictors for other types 
is harder as they lack the structure that is present in mathematical 
formulas.

\paragraph{Theorem List}
The type of theorem lists is the most commonly used type in \holfour proofs.
Here are two examples of such constructions with placeholder 
\texttt{[thm\_list]}: \texttt{REWRITE\_TAC [thm\_list]},  \texttt{METIS\_TAC 
[thm\_list]}.

In general, the list of arguments in such tactics respect the two following 
property:
\begin{itemize}
\item Additional theorems do not cause the tactic to fail and useless theorems 
are ignored. For \texttt{REWRITE\_TAC}, loops are generally avoided because 
most equations appears with only one orientation in the database of theorems.
\item The order of theorems in the list does not influence the final result and 
thus we can consider the list to be a set in our algorithm. This is 
particularly true for the first-order theorem prover \texttt{METIS\_TAC}.
\end{itemize}

Since our predictor is not perfectly accurate and the presented tactics are 
robust, we select a list of theorems usually longer that the original one to 
instantiate these tactics.
Our default predictor for list of theorems selects 16 theorems from the 
database of theorems $\mathbb{D}_{\mathit{theorem}}$. It uses the produced list
$\mathit{Predictions}^{\mathit{theorem}}_{16}(g)= [thm1,\ldots,thm16]$ to 
replace the
placeholders in $\hat{t}$. Thus, the instantiation of the first construction 
would be \texttt{REWRITE\_TAC [thm1,...,thm16]}.

\paragraph{Theorem}
To instantiate a single theorem correctly is much harder than finding a 
superset of theorems. The following examples with placeholder \texttt{thm} 
illustrate this point:
  \texttt{MATCH\_MP\_TAC thm}, \texttt{INDUCT\_THEN thm ASSUME\_TAC}
  
Here, the two examples accept only very specific kind of theorems: the theorem 
argument should match the precondition of the goal for \texttt{MATCH\_MP\_TAC} 
and an induction principle should be given to \texttt{INDUCT\_THEN}.
The likelihood of finding such theorem is low, thus given the list of 
predictions $\mathit{Predictions}^{\mathit{theorem}}_{16}(g)= 
[thm1,\ldots,thm16]$, we create 16 instantiations and try them in the 
order given by the predictor, and we return the result of the first successful 
tactic and produces new goals. With a modified version of 
the tactical \texttt{ORELSE} that fails if the tactic 
did not have any effect, we
can write for the first example a single new tactic that implements this 
procedure: 
  \[\texttt{MATCH\_MP\_TAC thm1 ORELSE ... ORELSE MATCH\_MP\_TAC thm16}\]

\paragraph{Term}
Some proofs require to instantiate a tactic argument by a term. Here are two 
abstracted 
tactics that use term 
arguments with placeholders \texttt{term} and \texttt{thm}: 
  \[\texttt{EXISTS\_TAC term},\ \texttt{SPEC\_THEN term STRUCT\_CASES\_TAC 
  thm}\]
Finding a witness to instantiate an existential formula is challenging because 
it is likely that this witness may not have been proposed before and there is 
no 
general rule to generate it. Instantiating 
the term argument in the second tactic is easier because the expected term is a 
variable of the goal. That is why we think that the first step 
toward term argument prediction is to select subterms of the goal ordered by 
their similarity with the term argument appearing in the original tactic. 
Additional experiments, using a tactic that tries the 16 closest subterms 
$[term1,\ldots,term16]$ to the goal with the trial and error procedure 
described in the previous paragraph, show that this simple term predictor 
improves the success 
rate of 
\tactictoe by another four percent. With the modified 
\texttt{ORELSE} tactical, the tactic derived from the abstraction
\texttt{EXISTS\_TAC term} in this experiment can be written as:
   \[\texttt{EXISTS\_TAC term1 ORELSE ... ORELSE EXISTS\_TAC term16}\]
    
In the future, with stronger machine learning models, we should be able to 
select the right term from larger sets of terms including:
\begin{itemize}
 \item subterms of the goal (already implemented),
 \item terms appearing as arguments of the abstracted tactic,
 \item tactic sub-expressions of type term and their subterms,
 \item subterms of theorems from the database $\mathbb{D}_{\mathit{theorem}}$,
 \item terms created by term synthesis techniques.
\end{itemize}

\begin{remark} (Limitations)
In the experiments presented in Section~\ref{sec:evaluation}, the 
abstraction/instantiation method is evaluated only for the type ``theorem 
list''.
Another issue is that the argument predictions are only influenced by the input 
goal. 
We believe that it would be beneficial to have
more flexible predictors that can adapt their selection to the abstracted 
tactic they procure instantiations for.
\end{remark}

\subsection{Selection of Abstracted Tactics}
As abstracted tactics do not appear in human proofs, we need to find a way to
predict them so that they can contribute to the \tactictoe proof search.
A straightforward idea is to try $\hat{t}$ before $t$ during the proof search.
However, this risks doing unnecessary work as the two may perform similar steps.
Therefore, we would like to decide
beforehand if one is better than the other.
In fact, we can re-use the orthogonalization module to do this task for us.
We add $\hat{t}$ to the competition, initially giving it the coverage of $t$.
If $\hat{t}$
wins, it is associated with $g$ and is included in the database of tactic
features and thus can be predicted during proof search.
After many orthogonalization competitions, the coverage of $\hat{t}$ may exceed
the coverage of
$t$. At
this point, the coverage of $\hat{t}$ is estimated on its own and not inherited
from
$t$ anymore.

\section{Proof Recording}\label{sec:recording}
Recording proofs in an LCF-style proof assistant can be done at different
levels.
In \holfour all existing approaches relied on modifying the kernel. This was
used
either to export the primitive inference
steps~\cite{Wong95recordingand,DBLP:conf/itp/KumarH12}
or to record dependencies of theorems~\cite{tgck-cpp15}. This was not suitable
for our
purpose of learning proving strategies at the intermediate tactic level. We
therefore
discuss recording proofs in an LCF-style proof assistant, with the focus on
\holfour
in this section.
Rather than relying on the underlying programming language, we parse the
original script file containing tactic proofs. This enables us to extract the
code of each tactic.
Working with the string representation of a tactic is better than working with
its value:
\begin{itemize}
\item Equality between two functions is easy to compute from
their string representation, which allows us to avoid
predicting the same tactic repeatedly.
\item
The
results of
\tactictoe can be returned in a readable form to the user. In this way, the
user can learn from the feedback, analyze the proof and possibly improve on it.
Furthermore \tactictoe does not need to be installed to run
tactic proofs generated by \tactictoe. In this way, the further development of
\tactictoe does not affect the robustness of \holfour.
\item
It is difficult (probably impossible) to transfer \sml values
between theories, since they are not run in the same \holfour session. In
contrast,
\sml code can be exported and imported.
\item To produce a tactic value from a code is easy in \sml, which can be
achieved by using a reference to a tactic and updating it with the function
\texttt{use}.
\end{itemize}

In order to transfer our tactic knowledge between theories, we want to be able
to re-use the code of a tactic recorded in one theory in another.
We also would like to make sure that the code is interpretable and that
its interpretation does not change in the other theory.

Even when developing one theory, the context is continually changing:
modules are opened and local identifiers are defined. Therefore, it is unlikely
that code extracted from a tactic proof is interpretable in any other part of
\holfour without any post-processing.
To solve this issue, we recursively replace each local identifier by its
definition until we are able to write any expression with global identifiers
only. Similarly, we prefix each global identifier by its module.
We call this process \emph{globalization}. The result is a standalone \sml code
executable in any \holfour theory.  In the absence of side effects, it is
interpreted in the same way across theories.
Therefore, we can guarantee that the behavior of recorded stateless tactics
does not change. Even with some stateful tactics, the prediction algorithm is
effective because most updates on the state of \holfour increase the
strength of stateful tactics.

\subsection{Implementation}
We describe in more detail our implementation of the recording algorithm. It
consists of 4 phases: tactic proof extraction, tactic proof globalization,
tactic unit wrapping, and creation of tactic-goal pairs.

Because of the large number of \sml constructions, we only describe the effect
of these steps on a running example that contains selected parts of the theory
of lists.

\begin{example}\label{ex:running}(Running example)
\small
\begin{lstlisting}[language=SMLSmall]
open boolLib Tactic Prim_rec Rewrite
ldots
val LIST_INDUCT_TAC = INDUCT_THEN list_INDUCT ASSUME_TAC
ldots
val MAP_APPEND = store_thm("MAP_APPEND",
  ``!(f:'a->'b) l1 l2. MAP f (APPEND l1 l2) = APPEND (MAP f l1) (MAP f l2)``,
  STRIP_TAC THEN LIST_INDUCT_TAC THEN ASM_REWRITE_TAC [MAP, APPEND])
\end{lstlisting}
\end{example}

The first line of this script file opens modules (called structures in \sml).
Each of
the modules contains a list of global identifiers which become directly
accessible in
the rest of the script.
A local identifier \texttt{LIST\_INDUCT\_TAC} is declared next, which is a
tactic that performs induction on lists. Below that, the theorem
\texttt{MAP\_APPEND} is proven.
The global tactic \texttt{STRIP\_TAC} first removes universal quantifiers. Then,
the goal is split into a base case and an inductive case. Finally, both
of these cases are solved by \texttt{ASM\_REWRITE\_TAC [MAP, APPEND]}, which
rewrites assumptions with the help of theorems previously declared in this
theory.

We first parse the script file to extract tactic proofs. Each of them is found
in
the third argument of a call to \texttt{store\_thm}. The result of tactic proof
extraction for the running example in presented in Example~\ref{ex:r0}.

\begin{example}\label{ex:r0}(Tactic proof extraction)
\small
\begin{lstlisting}[language=SMLSmall]
STRIP_TAC THEN LIST_INDUCT_TAC THEN ASM_REWRITE_TAC [MAP, APPEND]
\end{lstlisting}
\end{example}

In the next phase, we globalize identifiers of the tactic proofs.
Infix operators such as \texttt{THEN} need to be processed in a special way so
that they keep their infixity status after globalization. For simplicity,
the
globalization of infix operators is omitted in Example~\ref{ex:r1}.
In this example, the three main cases that can happen during the globalization
are depicted. The first one is the globalization of identifiers declared in
modules. The global identifiers \texttt{STRIP\_TAC} and
\texttt{ASM\_REWRITE\_TAC} are prefixed by their module \texttt{Tactic}. In
this way, they will still be interpretable whether \texttt{Tactic} was open or
not. The local identifier \texttt{LIST\_INDUCT\_TAC} is replaced by its
definition which happens to contain two global identifiers.
The previous paragraph describes the globalization for all identifiers except
local theorems.  We do not replace a local
theorem by its tactic proof (which its \sml definition) because we want to
avoid unfolding proofs inside other proofs.
If the theorem is stored in the \holfour database available across \holfour
developments, we can obtain the theorem value by calling \texttt{DB.fetch}.
Otherwise the globalization process fails and the local theorem identifier is
kept unchanged. A recorded tactic with an unchanged local theorem as argument
is only interpretable inside the current theory.

\begin{example}\label{ex:r1} (Globalization)
\begin{lstlisting}[language=SMLSmall]
Tactic.STRIP_TAC THEN
Prim_rec.INDUCT_THEN (DB.fetch "list" "list_INDUCT") Tactic.ASSUME_TAC THEN
Rewrite.ASM_REWRITE_TAC [DB.fetch "list" "MAP", DB.fetch "list" "APPEND"]
\end{lstlisting}
\end{example}

Running the globalized version of a tactic proof will have the exact same
effect as the original. But since we want to extract information from this
proof in the form of tactics and their input goals, we modify it further.
In particular, we need to define at which level we should record the tactics in
the tactic proof. The simplest idea would be to record all \sml subexpressions
of type \texttt{tactic}. However, it will damage the quality of our data by
dramatically increasing the number of tactics associated with a single goal.
Imagine a tactic proof of the form \texttt{A THEN B THEN C}; then the tactics
\texttt{A}, \texttt{A THEN B} and \texttt{A THEN B THEN C} would be valid
advice for something close to their common input goal. The tactic
\texttt{A THEN B THEN C} is likely to be too specific. In contrast, we can
consider
the tactic \texttt{REPEAT A} that is repetitively calling the tactic \texttt{A}
until
\texttt{A} has no effect. For such calls, it is often preferable to record
a single call to \texttt{REPEAT A} rather than multiple calls to \texttt{A}.
Otherwise branches after each call to \texttt{A} would be necessary as part
of proof search.
To sum up, we would like to record only the most general tactics which
make the most progress on a goal. As a trade-off between these two objectives,
we split proofs into tactic units.

\begin{definition}(Tactic unit)
A tactic unit is an \sml expression of type \texttt{tactic} that does not
contain an
infix operator at its root.
\end{definition}

Because such tactic units are constructed from visual information present in
the tactic proof, they often represent what a human user considers to be a
single step of
the proof.

To produce the final recording tactic proof, we encapsulate each tactic unit
in a recording function \texttt{R} (see Example~\ref{ex:wrap}). In order to
record
tactics in all \holfour theories, we replace the original proof by the
recording proof in each theory and rebuild the \holfour library.

\begin{example}\label{ex:wrap} (Tactic unit wrapping)
\begin{lstlisting}[language=SMLSmall]
R "Tactic.STRIP_TAC" THEN
R "Prim_rec.INDUCT_THEN (DB.fetch \"list\" \"INDUCT\") Tactic.ASSUME_TAC" THEN
R "Rewrite.ASM_REWRITE_TAC
  [fff \"list\" \"MAP\", fff \"list\" \"APPEND\"]"
\end{lstlisting}
\end{example}

At run time the function \texttt{R} is designed to observe what input goals a
tactic  receives without changing its output. The implementation of \texttt{R}
is presented in Example~\ref{ex:record}.

\begin{example}\label{ex:record} (Code of the recording function)
\begin{lstlisting}[language=SMLSmall]
fun R stac goal = (save (stac,goal); tactic_of_sml stac goal)
\end{lstlisting}
\end{example}

The function \texttt{save} writes the tactic-goal pair to disk increasing the
number of entries in our database of tactics. The function
\texttt{tactic\_of\_sml} interprets the \sml code \texttt{stac}. The tactic is
then applied to the input goal to replicate the original behavior.
After all modified theories are rebuilt, each call to a wrapped tactic in a
tactic proof is recorded as a pair containing its globalized code and its
input goal.

\section{Experimental Evaluation}\label{s:experiments}
The execution of \tactictoe's main loop in each re-proving experiment is
performed on a single CPU. An additional CPU is needed for experiments relying
on asynchronous \eprover calls.

\subsection{Methodology}
The evaluation imitates the construction of the library: For each theorem only
the previous human proofs are known. These are used as the learning base for
the predictions.
To achieve this scenario we re-prove all theorems during a modified build of
\holfour.
As theorems are proved, their tactical proofs and their statements are
recorded and included in the training examples.
For each theorem we first attempt to run the \tactictoe search with a time
limit of 60 seconds before processing the original tactic proof.
In this way, the fairness of the experiments is guaranteed by construction:
only previously declared \sml values (essentially tactics, theorems and
simpsets) are accessible to \tactictoe.

\paragraph{Datasets: optimization and validation}
All top-level theorems from the standard library are considered with the
exception of 440 hard problems (containing a \texttt{let} construction in their
proof) and 1175 easy problems (build from \texttt{save\_thm} calls).
Therefore, during the full-scale experiments, we evaluate 7164 theorems.
We use every tenth theorem of the first third of the standard library for
parameter optimization, which amounts to 273 theorems.

Although the evaluation of each set of parameters on its own is fair,
the selection of the best strategy in Section~\ref{sec:tuning} should also be
considered as a learning process. To ensure the global fairness, the final
experiment in Section~\ref{sec:full_exp} runs the best strategy on the full
dataset which is about 30 times larger.

\paragraph{Reproducibility}
The \tactictoe project is a part of the \holfour distribution 
which can be downloaded at \url{https://github.com/HOL-Theorem-Prover/HOL}.
The \tactictoe modules can be found in the directory
\url{HOL/src/tactictoe}. The \url{README} file explains how to launch the 
recording 
process (which includes learning techniques such as orthogonalization),
use \tactictoe interactively and run re-proving experiments on a theory. The 
recording process takes less than eight hours for the \holfour standard library. An upper bound for 
the time necessary for the final experiment can be given by the formula:
\[\mathit{number\ of\ theorems} \times \mathit{timeout} = 7164 \times 60\ 
\mathit{seconds} = 
4.975\ \mathit{days}\] 
This in practice translates 
to about $2.5\ \mathit{days}$, because successful searches stop before reaching 
the timeout. Finally, by running this experiment on 20 cores, we reduce
the actual wall clock time to several hours.

\paragraph{Portability}
Although \tactictoe is designed for \holfour, the \tactictoe modules 
implemented in this paper could be ported to another ITP. In order to 
facilitate this work, we give
ideas on how to adapt the major components of \tactictoe in the context of 
another ITP and warn about some of the issues that could arise.
First, the recording mechanism requires the ability to parse and manipulate 
proof scripts and include the possibility to define and recognize what a tactic 
is. In \holfour, \sml is our proof scripts language, so this was achieved by 
parsing the \sml programming language. In other ITPs, this task may be 
facilitated by the fact that proof scripts are written with a specifically 
designed language so the ability to modify syntax trees of proof scripts may 
already be available. Another issue that can negatively impact the quality of 
the recorded data is the fact that the effect of the tactics may depend on the 
state of the ITP (i.e. flags), which rarely happens in \holfour.
Porting the other modules (abstraction, orthogonalization, predictors and MCTS 
search) should be quite straightforward. Provided that application of 
tactics and the computation of predictions is fast enough (in the order of 
10 milliseconds each), the MCTS algorithm should explore enough of the tactic 
space to be a valuable tool for theorem proving.

\subsection{Tuning \tactictoe}\label{sec:tuning}
We optimize \tactictoe by tuning parameters of six different techniques:
the timeout of tactics, orthogonalization, abstraction, MCTS policy, MCTS
evaluation and ATP integration. During training, optional techniques such as
orthogonalization may also be turned off completely.
\tactictoe also includes other important techniques which are run with their
default parameters: feature extraction, prediction
algorithms, number of premises for ATPs and MCTS evaluation radius (set by default to
10).
By choosing a set of parameters, we create a strategy for
\tactictoe that is evaluated on our training set of 273 theorems. And the
strategy with the highest number of re-proven theorems is selected for
a full scale evaluation.

\begin{table}[ht]
\centering\ra{1.3}
\small
\begin{tabular}{llcc}
\toprule
 Technique & Parameters & Solved (1s) & Solved (60s)\\
\midrule
Tactic timeout & 0.02s && 154\\
                & 0.05s (default) && 156\\
                & 0.1s && 154\\
\midrule
Orthogonalization & none & 105 & 156 \\
                  & $radius = 10$ & 121 & 156 \\
                  & $radius = 20$ (default) & 121 & 156 \\
                  & $radius = 40$ & 124 & 156 \\
\midrule
Abstraction       & none (default)  && 156\\
                  & $theorems = 8$  && 195\\
                  & $theorems = 16$ && 199\\
                  & $theorems = 32$ && 195\\
\midrule
  MCTS policy & $c_{\mathit{policy}} = 0.4$ && 149\\
            & $c_{\mathit{policy}} = 0.5$ (default) && 156\\
            & $c_{\mathit{policy}} = 0.6$ && 153\\
\midrule
MCTS evaluation & none (default + best abstraction) && 199\\
        & $c_{\mathit{exploration}} = 1$ && 201\\
        & $c_{\mathit{exploration}} = 2$ && 203\\
        & $c_{\mathit{exploration}} = 4$ && 198\\
\midrule
ATP integration & none (default + best abstraction && 203\\
                &  + best MCTS evaluation) &&\\
                & \metis 0.1s && 216\\
                & \metis 0.2s && 212\\
                & \metis 0.4s && 212\\
                & \eprover && 213\\
                & \metis 0.1s + \eprover && 218\\
\bottomrule
\end{tabular}
\caption{\label{tab:tuning} Number of problem solved with different set of
parameters for \tactictoe on a training set of 273 theorems.}
\end{table}

In Table~\ref{tab:tuning}, the success rate of the different \tactictoe
strategies is presented.
The first four techniques are tested relative to the same baseline indicated in
the table by the tag ``default''.  This default strategy relies on a tactic
timeout of 0.05s, an orthogonalization radius of 20 and a policy coefficient of
0.5.
The two last techniques are tested relative
to a baseline consisting of the best set of parameters discovered so far and
are marked with the improvement over the ``default'' strategy.
In each subtable, each experiment differs from the current baseline by exactly
one parameter.

Thanks to the larger search time available in this experiment, the timeout for
each tactic can be increased from the 0.02 seconds used in \tactictoe's initial
experiments~\cite{tgckju-lpar17} to 0.05
seconds. Removing tactics with similar effect as performed by the
orthogonalization process is only beneficial when running \tactictoe for a short
period of time. It seems that the allotted time allows a strategy
running without orthogonalization to catch up. Yet, side experiments show
that orthogonalization becomes relevant again when we add additional
tactics through abstraction.
It is best to predict a list of 16 theorems to instantiate arguments of type
theorem list in tactics. At any rate, argument prediction for abstracted
tactics is the technique that has the highest impact on the success of
\tactictoe. Integration of ATPs being a specialization of this technique
contributes significantly as well. Experiments involving \eprover, run the ATP
using a separate process asynchronously with a timeout of 5 seconds and 128 premises.
The fact that \metis outperforms \eprover in this setting is due to the fact
that \metis is run with a very short timeout, allowing to close different part
of search tree quickly. Because \metis is weaker than \tactictoe as an ATP, it
is given less predictions. But even if an essential lemma for the proof of a
goal is not predicted, a modification on the goal performed by a tactic can
change its prediction to include the necessary lemma.
This effect minimizes the drawback of relying on a small number of predictions.

The MCTS evaluation, despite relying on the largest amount of collected data,
does not provide a significant improvement over a strategy relying on no
evaluation. The main reason is that our predictor learning abilities is
limited and we are using an estimate of the provability of lists of goals for
evaluation. A more accurate evaluation could be based on an estimation of the
length of the proof required to close a list of goals.

\subsection{Full-scale experiment}\label{sec:full_exp}

Based on the results of parameter tuning, we now evaluate a version of
\tactictoe with its best parameters on an evaluation set of 7164 theorems from
the \holfour standard library. We compare it with the performance of \eprover.
For reasons of fairness, \eprover asynchronous calls are not included in the
best \tactictoe strategy. The ATP \eprover is run in auto-schedule mode with
128 premises. The settings for \tactictoe are the following:
0.05 seconds tactic timeout, an orthogonalization radius of 20, theorem list
abstraction with 16 predicted theorems for instantiation, a prior policy
coefficient of 0.5, an evaluation radius of 10, an exploration coefficient of 2
and priority calls to \metis with a timeout of 0.1 seconds.

\begin{table}[h!]
\centering\ra{1.3}
\begin{tabular}{lc}
\toprule
  & Solved (60s) \\
\midrule
   \eprover   & 2472 (34.5\%)\\
   \tactictoe & 4760 (66.4\%)\\
\midrule
   Total  & 4946 (69.0\%)\\
\bottomrule
\end{tabular}
\caption{\label{tab:_param} Evaluation on 7164 top-level theorems of the
\holfour standard library
}
\end{table}

The results shows that \tactictoe is able to prove almost twice as many
theorems as \eprover. Combining the results of \tactictoe and \eprover we get a
69.0\% success rate which is significantly above the 50\% success rates
of hammers on this type of problems~\cite{tgck-cpp15}. Moreover, \tactictoe is
running a single set of parameters (strategy), whereas hammers and ATPs have
been optimized and rely on a wide range of strategies.

\paragraph{Reconstruction}
Tactic proofs produced by \tactictoe during this experiment are all verifiable
in \holfour. By the design of the proof search, reconstruction of \tactictoe
proof succeeds, unless one of the tactics modifies the state of \holfour in a
way that changes the behavior of a tactic used in the final proof. This has not
occurred a single time in our experiments. And a tactical proof generated by
\tactictoe during the final experiment takes on average 0.37 seconds to replay.
More details on how a proof is extracted from the search tree is given in
Section~\ref{sec:proofdisplay}.
Comparatively, we achieve a reconstruction rate of 95 percent for \eprover,
by calling \metis for 2.0 seconds with the set of theorems used
in \eprover's proof as argument.

\begin{table}[]
\centering
\setlength{\tabcolsep}{3mm}
\begin{tabular}{@{}ccccc@{}}
\toprule
\phantom{ab} & {arith} & {real} & {compl} & {meas} \\
\midrule
\tactictoe & 81.2 & 74.0 & 79.6 & 31.3\\
\eprover & 59.9 & 72.0 & 67.1 & 12.8\\
\midrule
\phantom{abc} & {proba} & {list} & {sort} & {f\_map} \\
\midrule
\tactictoe & 45.8 & 79.5 & 65.3 & 82.0 \\
\eprover & 24.1 & 26.5 & 15.8 & 24.7 \\
\bottomrule
\end{tabular}
\caption{\label{theories} Percentage (\%) of re-proved theorems in the theories
\texttt{arithmetic}, \texttt{real}, \texttt{complex}, \texttt{measure},
\texttt{probability}, \texttt{list}, \texttt{sorting} and \texttt{finite\_map}.
}
\end{table}

Table~\ref{theories} compares the re-proving success rates for different
\holfour theories. \tactictoe outperforms \eprover on every
considered theory.
\eprover is more suited to deal with dense theories such as
\texttt{real} or \texttt{complex} where a lot of related theorems are available
and most proofs are usually completed by rewriting tactics. Thanks to its
ability to re-use custom-built tactics, \tactictoe
largely surpasses \eprover on theories relying on inductive terms and
simplification sets such as \texttt{arithmetic}, \texttt{list}
and \texttt{f\_map}. Indeed, \tactictoe is able to recognize where and when to
apply induction, a task at which ATPs are known to struggle with.

\paragraph{Reinforcement learning}
All our feature vectors have been learned form human proofs. We can now
also add tactic-goal pairs that appears in the last proof of \tactictoe. To
prevent
duplication of effort, orthogonalization of those
tactics is essential to have a beneficial effect.
Since recording and re-proving are intertwined during evaluation, the
additional data is available for the next proof search.
The hope is that the algorithm will improve faster by learning from its own
discovered proofs than from the human-written proofs~\cite{DBLP:conf/cade/Urban07}. Side experiments show that
this one shot reinforcement learning method increases \tactictoe's success rate
by less than a percent.

\subsection{Complexity of the proof search}
\pgfplotscreateplotcyclelist{my black}{
solid, mark repeat=100, mark phase=0, black!100\\
dashed, mark repeat=100, mark phase=0, black!100\\}

\begin{figure}[h]
\centering
\begin{tikzpicture}[scale=1]
\begin{axis}[
  legend style={anchor=south east, at={(0.9,0.1)}},
  width=\textwidth,
  height=0.7*\textwidth,xmin=0, xmax=60,
  ymin=0, ymax=4800,
  xtick={},
  ytick={},
  cycle list name=my black]
\addplot table[x=time, y=solved] {data/tactictoe_time};
\addplot table[x=time, y=solved] {data/eprover_time};
\legend{\tactictoe,\eprover}
\end{axis}
\end{tikzpicture}
\caption{\label{fig:c1} Number of problems solved in less than x seconds.}
\end{figure}
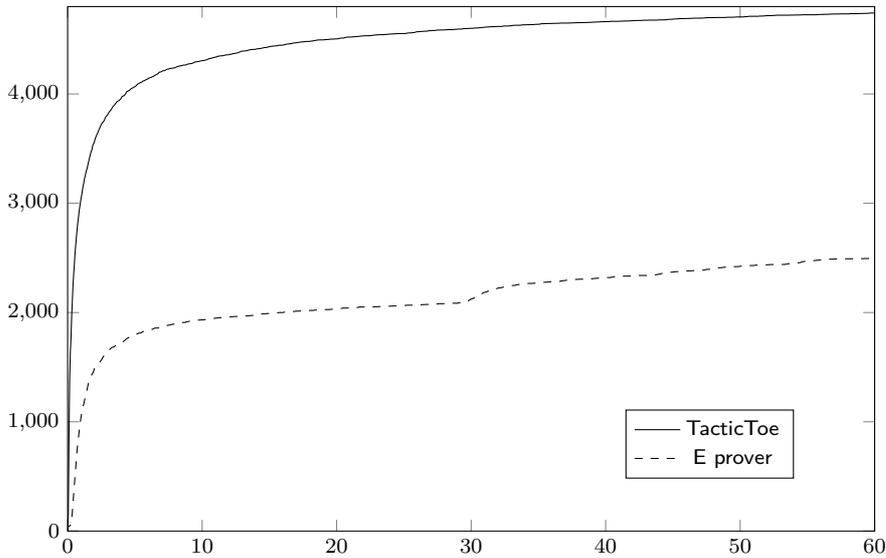

We first investigate how \tactictoe and \eprover scale as the search time
grows in Figure~\ref{fig:c1}.
In 10 seconds, \tactictoe solves 90 percent of the problems it can
solve in 60 seconds. The analysis is a bit different for \eprover. Indeed,
we can clearly deduce from the bump at 30 seconds that \eprover is using at
least two strategies and probably more. Strategies are a useful method to
fight the exponential decline in the number of newly proven theorems over
time. Therefore, integrating strategy scheduling in \tactictoe could be
something to be experimented with.

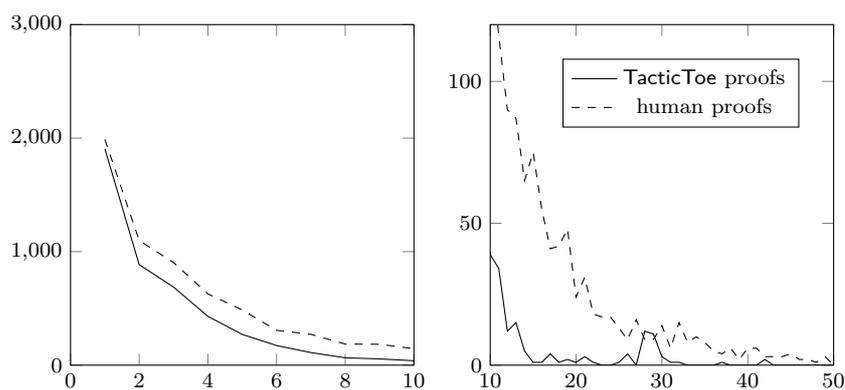
\begin{figure}[h]
\centering
\begin{tikzpicture}[scale=1]
\begin{axis}[
  legend style={anchor=north east, at={(0.9,0.9)}},
  width=0.5*\textwidth,
  height=0.5*\textwidth,
  ymin=0, ymax=3000,
  xmin=0, xmax=10,
  xtick={},
  ytick={},
  cycle list name=my black]
\addplot table[x=length, y=proofs] {data/tactictoe_proof_length};
\addplot table[x=length, y=proofs] {data/original_proof_length};
\end{axis}
\end{tikzpicture}
\begin{tikzpicture}[scale=1]
\begin{axis}[
  legend style={anchor=north east, at={(0.9,0.9)}},
  width=0.5*\textwidth,
  height=0.5*\textwidth,
  ymin=0, ymax=120,
  xmin=10, xmax=50,
  xtick={},
  ytick={},
  cycle list name=my black]
\addplot table[x=length, y=proofs] {data/tactictoe_proof_length};
\addplot table[x=length, y=proofs] {data/original_proof_length};
\legend{\tactictoe proofs,human proofs}
\end{axis}
\end{tikzpicture}
\caption{\label{fig:distrib}Number of tactic proofs with $x$ tactic units.}
\end{figure}

In order to appreciate the difficulty of the evaluation set from a human
perspective, the length distribution of human proofs in the
validation set is shown in Figure~\ref{fig:distrib}.  It is clear from the
graph that most of the human proofs are short. The proofs found by \tactictoe
follow a similar
distribution.
If \tactictoe finds a proof, there is about a 50
percent chance that it will be shorter than what a human would come up with.

\begin{figure}[h]
\centering
\begin{tikzpicture}[scale=1]
\begin{axis}[
  legend style={anchor=north east, at={(0.9,0.9)}},
  width=\textwidth,
  height=0.7*\textwidth,xmin=0, xmax=20,
  ymin=0, ymax=100,
  xtick={},
  ytick={},
  cycle list name=my black]
\addplot table[x=oplen, y=solved] {data/tactictoe_by_oplen};
\addplot table[x=oplen, y=solved] {data/eprover_by_oplen};
\legend{\tactictoe,\eprover}
\end{axis}
\end{tikzpicture}
\caption{\label{fig:percentage}Percentage of problem solved with respect to the
length of the
original proof until length 20.}
\end{figure}
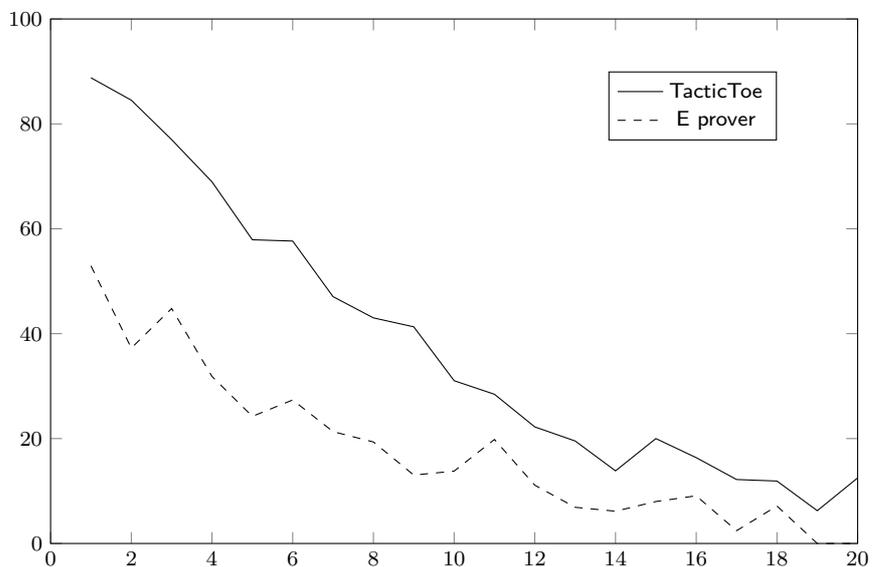

In Figure~\ref{fig:percentage}, we regroup problems by the length of their
human proof to measure how well \tactictoe and \eprover cope with increasing
levels
of difficulty. As expected, the longer the original proof is, the harder it is
for \tactictoe and \eprover to re-prove the theorem on their own.
The performance of \tactictoe is compelling with more than half of the theorems
that required a proof of length six being re-proven. It is also consistently
better than \eprover for any proof length.

\section{Minimization and Embellishment}\label{sec:proofdisplay}
When \tactictoe finds a proof of an initial goal, it returns a search tree
$\mathfrak{T}$ where the root node containing the initial goal is solved.
In order to transform this search tree into a tactic proof, we need to extract
the tactics that contributed to the proof and combine them using tactic
combinators.

By the design of the search, a single tactic combinator, \texttt{THENL}, is
sufficient. It combines a tactic $t$ with a list of subsequent ones, in such a
way that after $t$ is called, for each created goal a respective
tactic from the list is called.

Let $T_{sol}$ be a partial function that from a goal $g$ returns a tactic $t$
for which $A(g,t)$ is solved in $\mathfrak{T}$.
The proof extraction mechanism is defined by mutual
recursion on goals and nodes of $\mathfrak{T}$ by the respective function
$P_{goal}$ and $P_{node}$:

\begin{align*}
P_{goal}(g) &=_{\mathrm{def}} T_{sol}(g)\ \texttt{THENL}\ P_{node}(A(g,T_{sol}(g)))\\
P_{node}(a) &=_{\mathrm{def}} [P_{goal}(g_1),\ldots,P_{goal}(g_n)]\ \ \ \text{with}\
G(a) = g_1,\ldots,g_n\\
\end{align*}

The extracted tactic proof of $\mathfrak{T}$ is $P_{goal}(g_{root})$.
We minimally improve it by substituting \texttt{THENL} by \texttt{THEN} when the
list of goals is a singleton and removing \texttt{THENL []} during the
extraction phase.
Further post-processing such as
eliminating unnecessary tactics and theorems has been developed and
improves the user experience greatly~\cite{DBLP:conf/sefm/Adams15}.

\paragraph{Proof Length Minimization}
A fast and simple minimization is applied when processing the final proof. If
the tactic \texttt{A THEN B} appears in the proof and has the same effect as
\texttt{B} then we can replace \texttt{A THEN B} by \texttt{B} in the proof.
Optionally, stronger minimization can be obtained
by rerunning \tactictoe with a low prior
policy coefficient, no prior evaluation and a database of tactics that contains
tactics from the discovered proof only.

\paragraph{Tactic Arguments Minimization}
Let $t$ be a tactic applied to a goal $g$ containing a list $l$ (of theorems)
as argument. And let $t'$ be the tactic $t$ where one element $e$ of $l$ as be
removed. If $t$ and $t'$ have the same effect on $g$ then $t'$ can replace $t$
in the final proof. This process is repeated for each element of $l$.
This is a generalization of the simplest method used for minimizing a list of
theorems in ``hammers'' \cite{hammers4qed}.

\paragraph{Embellishment}
Without embellishment, the returned tactic is barely readable as it contains
information to guarantee that each \sml subterm is interpreted in the same way 
in any context. Since we return the proof at a point where the \sml interpreter 
is
in a specific state, we can strip unnecessary information, such as
module prefixes. If possible, we group all local declarations in the
proof under a single \texttt{let} binding at the start of the proof. We
also replace extended terms by their quoted version.
All in all, if a prettified tactic $t_p$ has the same effect its original
$t_o$, we replace $t_p$ by $t_o$ in the proof.\\

The total effect of these minimization and embellishment techniques
on the readability of a tactic proof is depicted in
Example~\ref{ex:pretty}.

\begin{example}(Theorem \texttt{EVERY\_MAP})\label{ex:pretty}\\\
Before minimization and embellishment:
\begin{lstlisting}[language=SMLSmall]
boolLib.REWRITE_TAC [DB.fetch "list" "EVERY_CONJ", DB.fetch "list" "EVERY_MEM",
  DB.fetch "list" "EVERY_EL", ldots ,  combinTheory.o_DEF] THEN
BasicProvers.Induct_on [HolKernel.QUOTE " (*#loc 1 11380*)l"] THENL
  [BasicProvers.SRW_TAC [] [],
   simpLib.ASM_SIMP_TAC (BasicProvers.srw_ss ()) [boolLib.DISJ_IMP_THM,
     DB.fetch "list" "MAP", DB.fetch "list" "CONS_11", boolLib.FORALL_AND_THM]]

\end{lstlisting}
After minimization and embellishment:
\begin{lstlisting}[language=SMLSmall]
Induct_on `l` THENL
  [SRW_TAC [] [], ASM_SIMP_TAC (srw_ss ()) [DISJ_IMP_THM, FORALL_AND_THM]]
\end{lstlisting}
\end{example}

\section{Case Study}\label{sec:case_study}
Since the proofs generated by \tactictoe are meant to be parts of a \holfour 
development, it is interesting to compare their quality with the original human 
proofs.
The quality of a proof in \holfour can be measured in terms of length, readability, maintainability and verification speed.
We study these properties in three examples taken from our full-scale experiment.
We list the time needed by \holfour to check a proof in parentheses.

We start with Example~\ref{ex:cs1}, which proves that greatest common divisors are unique.
The human formalizer recognized that it follows from two theorems.
The relevance filtering of \tactictoe is not as powerful as that used in hammers
and is therefore not able to find the \texttt{DIVIDES\_ANTISYM} property. The proof
proceeds instead by rewriting the
goal with the definitions then splitting the goal into multiple cases until the goal obligations are similar
to \texttt{DIVIDES\_ANTISYM}. As expected, \tactictoe's proof
also takes much longer to check.

\begin{example}\label{ex:cs1} \texttt{IS\_GCD\_UNIQUE} in theory \texttt{gcd}
\begin{lstlisting}[language=SMLSmall]
!a b c d. is_gcd a b c wedge is_gcd a b d ==> (c = d)
\end{lstlisting}
Human proof (5 milliseconds)
\begin{lstlisting}[language=SMLSmall]
PROVE_TAC[IS_GCD, DIVIDES_ANTISYM]
\end{lstlisting}
\tactictoe proof (80 milliseconds)
\begin{lstlisting}[language=SMLSmall]
STRIP_TAC THEN
REWRITE_TAC [fetch "gcd" "is_gcd_def"] THEN
REPEAT Cases THENL
  [METIS_TAC [],
   REWRITE_TAC [SUC_NOT, ALL_DIVIDES_0, compute_divides] THEN
     METIS_TAC [NOT_SUC],
   METIS_TAC [NOT_SUC, DIVIDES_ANTISYM],
   METIS_TAC [LESS_EQUAL_ANTISYM, DIVIDES_LE, LESS_0],
   METIS_TAC [],
   RW_TAC numLib.arith_ss [divides_def],
   METIS_TAC [DIVIDES_ANTISYM],
   METIS_TAC [LESS_EQUAL_ANTISYM, DIVIDES_LE, LESS_0]]
\end{lstlisting}
\end{example}

In Example~\ref{ex:cs2}, we try to prove that for any surjective function $f$
from $s$ to $t$, there exists an injective function $g$ from $t$ to $s$ such
that $f\ o\ g$ is the identity function. The human proof is quite complicated.
In contrast, \tactictoe finds a much smaller proof that expands the
definition of injectivity and surjectivity and calls \metis.
However, the \tactictoe's proof takes much longer to check due to the proof search happening inside \metis.

\begin{example}\label{ex:cs2} \texttt{SURJ\_INJ\_INV} in theory
\texttt{pred\_set}
\begin{lstlisting}[language=SMLSmall]
!f s t. SURJ f s t ==> ?g. INJ g t s wedge !y. y IN t ==> (f (g y) = y)
\end{lstlisting}
Human proof (2 milliseconds)
\begin{lstlisting}[language=SMLSmall]
REWRITE_TAC [IMAGE_SURJ] THEN
DISCH_TAC THEN Q.EXISTS_TAC `THE o LINV_OPT f s` THEN
BasicProvers.VAR_EQ_TAC THEN REPEAT STRIP_TAC THENL
  [irule INJ_COMPOSE THEN Q.EXISTS_TAC `IMAGE SOME s` THEN
     REWRITE_TAC [INJ_LINV_OPT_IMAGE] THEN REWRITE_TAC [INJ_DEF, IN_IMAGE] THEN
     REPEAT STRIP_TAC THEN REPEAT BasicProvers.VAR_EQ_TAC THEN
     FULL_SIMP_TAC std_ss [THE_DEF],
   ASM_REWRITE_TAC [LINV_OPT_def, o_THM, THE_DEF] THEN
     RULE_ASSUM_TAC (Ho_Rewrite.REWRITE_RULE
       [IN_IMAGE', GSYM SELECT_THM, BETA_THM]) THEN ASM_REWRITE_TAC []]
\end{lstlisting}
\tactictoe proof (50 milliseconds)
\begin{lstlisting}[language=SMLSmall]
SRW_TAC [] [SURJ_DEF, INJ_DEF] THEN METIS_TAC []
\end{lstlisting}
\end{example}

In Example~\ref{ex:cs3}, we prove a theorem about lists, a domain where \tactictoe excels compared to ATPs.
Given two list $l_1$ and $l_2$ where the length of $l_1$ (denoted $p$) is less
than a natural
number $n$, the theorem states updating the $n^{th}$
element of the concatenation of $l_1$ and $l_2$ is the same as
updating the $m^{th}$ element of $l_2$ where $m = n - p$. Again, the
\tactictoe's proof is
much more readable. It starts by applying induction on $l$. It solves the base
case of the induction by rewriting and proceeds by cases on $n=0$ or $n>0$
in the induction hypothesis. It finalizes the proof with a short call to \metis
using the definition of
\texttt{LUPDATE}. Here, \tactictoe's proof is arguably smaller, faster and
easier to
understand and maintain. Such proofs after an expert review could replace the
their respective original human proofs in the \holfour repository.

\begin{example}\label{ex:cs3} \texttt{LUPDATE\_APPEND2} in theory
\texttt{rich\_list}
\begin{lstlisting}[language=SMLSmall]
!l1 l2 n x. LENGTH l1 <= n ==>
  (LUPDATE x n (l1 ++ l2) = l1 ++ (LUPDATE x (n - LENGTH l1) l2))
\end{lstlisting}
Human proof (63 milliseconds)
\begin{lstlisting}[language=SMLSmall]
  rw[] THEN simp[LIST_EQ_REWRITE] THEN Q.X_GEN_TAC `z` THEN
  simp[EL_LUPDATE] THEN rw[] THEN simp[EL_APPEND2,EL_LUPDATE] THEN
  fs[] THEN Cases_on `z < LENGTH l1` THEN
  fs[] THEN simp[EL_APPEND1,EL_APPEND2,EL_LUPDATE]
\end{lstlisting}
\tactictoe proof (17 milliseconds)
\begin{lstlisting}[language=SMLSmall]
Induct_on `l1` THENL [SRW_TAC [] [],
  Cases_on `n` THENL [SRW_TAC [] [],
    FULL_SIMP_TAC (srw_ss ()) [] THEN METIS_TAC [LUPDATE_def]]]
\end{lstlisting}
\end{example}

In Example~\ref{ex:cs4}, we experiment with \tactictoe on a
goal that does not originate from the \holfour library. The conjecture is
that the set of
numbers $\{0,...,n+m-1\} \setminus \{0,...,n-1\}$ is the same as the set
obtained by adding $n$ to everything in $\{0,...,m-1\}$. In this example,
\tactictoe uses the simplification set \texttt{ARITH\_ss} to reduce arithmetic
formulas. This exemplifies another advantage that \tactictoe has over ATPs,
namely its ability to take advantage of user-defined simplification sets.

\begin{example}\label{ex:cs4}
\begin{lstlisting}[language=SMLSmall]
count (n+m) DIFF count n = IMAGE ((+) n) (count m)
SRW_TAC [ARITH_ss] [EXTENSION, EQ_IMP_THM] THEN
Q.EXISTS_TAC `x - n` THEN
SRW_TAC [ARITH_ss] []
\end{lstlisting}
\end{example}

\section{Related Work}
There are several essential components of our work that are comparable to
previous approaches: tactic-level proof recording, tactic
selection through machine learning techniques and automatic tactic-based proof
search. Our work is also related to previous approaches that use machine
learning to select premises for the ATP systems and guide ATP proof search
internally.

In \hollight, the Tactician tool~\cite{DBLP:conf/sefm/Adams15}
can transform a packed tactical proof into a series of interactive tactic
calls. Its principal application
was so far refactoring the library and teaching common proof techniques to new
ITP users. In our work, the splitting of a proof into a sequence of tactics is
essential for the
tactic recording procedure, used to train our tactic prediction mechanism.

\textsf{SEPIA}~\cite{DBLP:conf/cade/GransdenWR15} can
generate
proof scripts from previous \coq proof examples.
Its strength lies in the ability to produce probable sequences
of tactics for solving domain specific goals. It operates by creating a model
for common sequences of tactics for a specific library.
This means that in order to propose the next tactic, only the previously
called tactics
are considered.
Our algorithm, on the other hand, relies mainly on the characteristics of the
current goal
to decide
which tactics to apply next. In this way, our learning mechanism has to
rediscover why each
tactic was applied for the current subgoals. It may lack some useful bias for
common sequences
of tactics, but is more reactive to subtle changes. Indeed, it can be trained
on a large library and only tactics relevant to the current subgoal will be
selected.
Concerning the proof search, \textsf{SEPIA}'s 
breadth-first search is replaced by MCTS which allows for supervised learning
guidance in the exploration of the search tree.
Finally, \textsf{SEPIA} was evaluated on three chosen parts of the
\coq library demonstrating that it globally outperforms individual \coq
tactics. Here, we demonstrate the competitiveness of our system against
\eprover on the \holfour standard library.

\textsf{ML4PG}~\cite{DBLP:journals/corr/abs-1212-3618,DBLP:journals/mics/HerasK14}
groups related proofs using clustering
algorithms. It allows \coq users to inspire themselves from similar proofs and
notice duplicated proofs. Comparatively, our predictions come from a much more
detailed description of the target goal. \tactictoe can also organize the
predictions to produce verifiable proofs and is not restricted to user
advice.

A closely related approach is proof planning that was first described in 
~\cite{DBLP:conf/cade/Bundy88}. Proof planners have been implemented to 
facilitate interactive theorem proving 
development in Oyster~\cite{DBLP:conf/cade/BundyHHS90},  
Omega~\cite{DBLP:conf/cade/BenzmullerCFFHKKKMMSSS97} and 
\isabelle~\cite{DBLP:conf/cade/DixonF03}.
A proof plan can be seen as a program that given a goal 
returns the appropriate tactic for this goal. From this point of view, 
\tactictoe is a general proof plan that learns to apply the right tactic from 
examples and has a backtracking mechanism given by the MCTS algorithm.
The philosophy of proof planning is also to create a few 
tactics that can prove a wide range of theorems. 
Apart from the 
abstraction technique, \tactictoe is not yet synthesizing any tactics but simply 
selecting them out of a large tactic set. This may become necessary in the future, 
using a variety of statistical/symbolic learning methods.
More general tactics are usually easier to predict 
accurately by machine learning models as they are associated with more 
(successful) examples in our database.

Proof patching \cite{RingerYLG18} is an approach that attempts to learn the
proof changes necessary for a modification of a prover library or system. This
approach is however hard to use to create completely new proofs. Following this idea, \tactictoe
could try to learn from different versions of the prover library.

It is important to consider the strength of the language used for building 
proofs and proof procedures when designing a system such as \tactictoe. Indeed, 
such system will be re-using those procedures.
In \isabelle, the language {\textsf{Eisbach}\xspace}~\cite{eisbach} was created 
on top of the standard tactic language by giving access to constructors of the 
tactic language to the end-user. The language 
{\textsf{PSL}\xspace}~\cite{NagashimaK17psl} was also developed. It can combine 
regular tactics and tools such as \sledgehammer.
In \coq, the \ltac~\cite{ltac} meta language was designed to enrich the 
tactic combinators of the \coq language. It adds functionalities such as 
recursors and matching operators for terms and proof contexts.
In \holfour, the strength of \sml as a proof strategy 
language has been demonstrated by the implementation of complex proof 
procedures such as \metis. 

Machine learning has also been used to advise the best library lemmas for new
ITP goals.
This can be done either in an interactive way, when the user completes the
proof based on the recommended lemmas, as in the Mizar Proof Advisor~\cite{Urb04-MPTP0}, or attempted fully automatically, where such lemma
selection is handed over to the ATP component of a \emph{hammer}
system, as in the hammers for \holfour~\cite{tgck-cpp15}, 
\hollight~\cite{holyhammer}, \isabelle~\cite{BlanchetteGKKU16} and \mizar~\cite{UrbanRS13,mizAR40}.

Internal learning-based selection of tactical steps inside an ITP is analogous
to internal learning-based selection of clausal steps inside ATPs such as
\textsc{MaLeCoP}~\cite{malecop} and \textsc{FEMaLeCoP}~\cite{femalecop}. These
systems
use the naive Bayes classifier to  select clauses for the extension steps in
tableaux proof search based on many previous proofs. Satallax~\cite{Brown2012a}
can guide its
search internally~\cite{mllax} using a command classifier, which can estimate
the priority of the 11 kinds of
commands in the priority queue based on positive and negative examples.

\section{Conclusion}\label{sec:concl}
We proposed a new proof assistant automation technique which combines
tactic-based proof search with machine learning prediction.
Its implementation, \tactictoe, achieves an overall success rate of 66.4\%
on 7164 theorems of the \holfour standard library, surpassing \eprover
with auto-schedule. Its
effectiveness is especially visible on
theories which use inductive data structures, specialized decision procedures,
and custom built simplification sets.
Thanks to the learning abilities of \tactictoe, the generated tactic proofs
often reveal the high-level structure of the proof. 
We therefore believe that predicting ITP tactics based on the current goal
features is a very reasonable approach to automatically guiding proof search,
and that accurate predictions can be obtained by learning from the knowledge
available in today's large formal proof corpora.

To improve the quality of the predicted tactics,
we would like to predict other type of arguments independently, as it was done
for theorems. In this direction, the most interesting arguments to
predict next are terms as they are ubiquitous in tactics. Further along the way,
new tactics could be created by programming them with any construction
available in \sml.
The proof search guidance can also be improved, for example by considering
stronger machine learning algorithms
such as deep neural networks as a model for the policy and evaluation in MCTS.
The quality of such models could be enhanced by
training and testing existing and synthesized tactics.
Conjecturing suitable intermediate steps could also allow \tactictoe to solve
problems which require long proofs by decomposing it into multiple easier
steps.

\paragraph{Acknowledgments}\label{sect:acks}
We would like to thank Lasse Blaauwbroek and Yutaka Nagashima for their
insightful comments which
contributed to improve the quality of this paper. This work has been supported
by the ERC Consolidator grant no.\ 649043 \textit{AI4REASON}, the ERC starting
grant no.\ 714034 \textit{SMART}, the Czech
project AI \& Reasoning CZ.02.1.01/0.0/0.0/15\_003/0000466 and the
European Regional Development Fund.

\bibliographystyle{plain}
\bibliography{biblio}

\end{document}